\documentclass{article}

\usepackage[francais,english]{babel} %
\usepackage[T1]{fontenc} %
\usepackage[utf8x]{inputenc} %
\usepackage{a4wide} %
\usepackage{palatino} %

\let\bfseriesbis=\bfseries \def\bfseries{\sffamily\bfseriesbis}

\usepackage[a4paper,top=2cm,bottom=2cm,left=2cm,right=2cm,marginparwidth=1.75cm]{geometry}

\usepackage{amsmath}
\usepackage{graphicx}
\usepackage{bussproofs} 
\usepackage[algoruled,french,onelanguage,linesnumbered]{algorithm2e} 

\usepackage[dvipsnames]{xcolor}
\definecolor{gris125}{gray}{0.875}
\definecolor{gris25}{gray}{0.75}
\definecolor{gris5}{gray}{0.5}
\definecolor{blanc}{gray}{0}

\definecolor{pale}{HTML}{F3EEDC}
\definecolor{bleu}{HTML}{B0C3CD}
\definecolor{rouge}{HTML}{EEC6C6}
\definecolor{coeurplus}{HTML}{EECBAA}
\definecolor{autre}{HTML}{606E5F}

\usepackage{lmodern}

\def \epsilon{\varepsilon}
\def \phi{\varphi}

\usepackage{amsfonts, amssymb, mathrsfs, esint}
\usepackage{amsthm}
\usepackage{wasysym}
\usepackage{stmaryrd}
\usepackage{verbatim}
\usepackage{tikz,tkz-tab,tkz-fct}
\usepackage{pstricks,pst-node,pst-tree,pst-coil}
\usepackage{pgf}
\usetikzlibrary{arrows,automata,shapes}

\usepackage{booktabs}

\usepackage{fancybox}
\usepackage{fancyhdr,lastpage,mdwlist}
\usepackage{enumerate}
\usepackage{multicol}
\usepackage{manfnt}
\usepackage{chemfig}
\usepackage{tabto}
\usepackage[inline]{asymptote}
\usepackage{wrapfig}

\usepackage{hyperref}

\usepackage{tcolorbox}
\tcbuselibrary{theorems}
\usepackage{varwidth}
\tcbuselibrary{skins}

\newtcbtheorem[auto counter, number within = section]{theo}{Théorème}
{enhanced,frame empty,interior empty,colframe=rouge,
	coltitle=black,fonttitle=\bfseries,colbacktitle=rouge!15!white,
	borderline={0.5mm}{0mm}{rouge},
	attach boxed title to top left={yshift=-2mm, xshift=10mm},
	boxed title style={boxrule=0.4pt},varwidth boxed title}{theo}

\newtcbtheorem[auto counter, number within = section]{propo}{Proposition}
{enhanced,frame empty,interior empty,colframe=rouge,
	coltitle=black,fonttitle=\bfseries,colbacktitle=rouge!15!white,
	borderline={0.5mm}{0mm}{rouge},
	attach boxed title to top left={yshift=-2mm, xshift=10mm},
	boxed title style={boxrule=0.4pt},varwidth boxed title}{theo}

\newtcbtheorem[auto counter, number within = section]{defi}{Définition}
{enhanced,frame empty,interior empty,colframe=bleu,
	coltitle=black,fonttitle=\bfseries,colbacktitle=bleu!15!white,
	borderline={0.5mm}{0mm}{bleu},
	attach boxed title to top left={yshift=-2mm, xshift=10mm},
	boxed title style={boxrule=0.4pt},varwidth boxed title
    }{theo}

\newtcbtheorem[no counter]{coeur}{À retenir $\heartsuit$}
{enhanced,frame empty,interior empty,colframe=coeurplus,
	coltitle=black,fonttitle=\bfseries,colbacktitle=coeurplus!15!white,
	borderline={0.5mm}{0mm}{coeurplus},
	attach boxed title to top left={yshift=-2mm, xshift=10mm},
	boxed title style={boxrule=0.4pt},varwidth boxed title}{theo}

\theoremstyle{definition}
\newtheorem{prop}{Proposition}

\newtheorem{app}{Application}
\newtheorem{theor}{Théorème}
\newtheorem{lem}{Lemme}

\usepackage{enumerate}

\newcommand{\cB}[1]{\textcolor{NavyBlue}{\textbf{\textit{#1}}}}
\newcommand{\cG}[1]{\textcolor{PineGreen}{\textbf{\textit{#1}}}}

\newif\ifinnerapp \newif\ifinnerabs

\def\uabs#1#2%
{\ifinnerabs%
(\lambda #1 . \,%
\innerappfalse\innerabsfalse #2)%
\else%
\lambda #1 . \,%
\innerappfalse\innerabsfalse #2%
\fi\innerappfalse\innerabsfalse}

\def\app#1#2%
{\ifinnerapp%
(\innerappfalse\innerabstrue #1 \, \innerapptrue\innerabstrue #2)%
\else%
\innerappfalse\innerabstrue #1 \, \innerapptrue\innerabstrue #2%
\fi\innerappfalse\innerabsfalse}

\newcommand\lex[1][]{\mathscr{L}_{\text{\scriptsize{#1}}}}

\newcommand\simples[1]{\Sigma_{\mathrm{SimpleSyn}(G)}}
\newcommand\logic[1]{\Sigma_{\mathrm{Log}}}

\newcommand\lexsimple[1]{\lex{}_{\mathrm{Syn}}}
\newcommand\lexlog[1]{\lex{}_{\mathrm{Log}}}

\newcommand{\m}[1]{\mathcal{#1}}
\renewcommand{\Tp}{\mathrm{Tp}}
\DeclareMathOperator{\ord}{ord}

\renewcommand{\b}{$\backslash$}
\newcommand{\ch}[1]{\textbf{<}\!#1\!\textbf{>}}

\newenvironment{point}[1]%
{\subsection*{#1}}%
{}

\usepackage{todonotes}

\begin{document}

\title{Traduction des Grammaires Catégorielles de Lambek dans les Grammaires Catégorielles Abstraites}

\author{Valentin D. Richard\\
{\tt valentin.richard\at ens-paris-saclay.fr}\\
sous la direction de Philippe de Groote\\ équipe SEMAGRAMME\\ Loria (Nancy)}

\date{du 11 juin au 27 juillet 2018}

\maketitle

\pagestyle{plain} %
\thispagestyle{empty}

\selectlanguage{english}

\begin{abstract}
    \centerline{\large Translation of Lambek Grammars into Abstract Categorial Grammars}\vspace{0.8cm}
    
Lambek Grammars (LG) are a computational modelling of natural language, based on non-commutative compositional types. It has been widely studied, especially for languages where the syntax plays a major role (like English). The goal of this internship report is to demonstrate that every Lambek Grammar can be, not entirely but efficiently, expressed in Abstract Categorial Grammars (ACG). The latter is a novel modelling based on higher-order signature homomorphisms (using $\lambda$-calculus), aiming at uniting the currently used models. The main idea is to transform the type rewriting system of LGs into that of Context-Free Grammars (CFG) by erasing introduction and elimination rules and generating enough axioms so that the cut rule suffices. This iterative approach preserves the derivations and enables us to stop the possible infinite generative process at any step. Although the underlying algorithm was not fully implemented, this proof provides another argument in favour of the relevance of ACGs in Natural Language Processing.
\end{abstract}

\selectlanguage{francais}

\section*{Fiche de synthèse}
\begin{point}{Le contexte général}

J'ai effectué mon stage de L3 dans le domaine de la linguistique informatique, un champ multidisciplinaire de la recherche qui vise à étudier les langues naturelles avec les formalismes informatiques. Plus précisément, le traitement automatique des langues (TAL) cherche à créer des outils de traitement de la langue naturelle écrite ou orale pour diverses applications (recherche d'information, traduction automatique,...). J'étais plus axé sur une approche théorique, ayant pour objectif de produire des modèles appuyés sur la logique (grammaire, syntaxe, sémantique,...). Le but est de trouver un formalisme complet et intéressant en pratique qui permette de rendre compte de la structure (et éventuellement sémantique) d'une phrase.

Les grammaires (catégorielles) de Lambek \cite{Lam58}, utilisées depuis longtemps, misent sur la syntaxe en modélisant les mots par des fonctions et arguments (ou plutôt leur type respectif). Ce symbolisme intéressant a donné lieu a de nombreuses variations qui s'ajustent mieux aux particularités de la langue naturelle. Cependant, peu de programmes ont été développés pour les simples grammaires de Lambek.

Les grammaires catégorielles abstraites (ACG) sont un formalisme créé par mon maître de stage Philippe de Groote \cite{dG01}. Elles ont l'avantage de modéliser la structure sous-jacente d'une phrase (les connections entre les mots) avec le même outil logique que le rendu final (la suite de mots). Dans l'optique de montrer que ce modèle est pertinent, les recherches notamment de l'équipe SEMAGRAMME s'orientent vers le développement d'outils pour les ACG  (\href{https://gitlab.inria.fr/ACG/dev/ACGtk}{ACGtk}) et la mise en relation de ce formalisme avec ceux déjà existants.

La question est donc de savoir s'il est possible d'exprimer, et comment, une grammaire de Lambek en ACG. Cela argumenterait en la faveur de ce dernier modèle pour son utilisation concrète dans des outils de TAL. Grâce aux travaux de Pentus \cite{Pen97}, des avancées ont déjà été faites par de Groote \cite{dG16}.

\end{point}

\begin{point}{Le problème étudié}

Insatisfait de sa première traduction, de Groote a eu l'idée d'une autre méthode, qu'il juge plus adéquate et dont il aimerait montrer la validité. L'objet de mon stage est donc de démontrer que cet autre procédé est correct. L'intérêt de ceci est de fournir une transformation plus cohérente linguistiquement, et un peu moins complexe algorithmiquement, donc plus pratique pour une potentielle réutilisation concrète dans un outil logiciel.

Le problème était le plus enclin être posé par l'équipe SEMAGRAMME (et particulièrement de Groote) puisque les ACG sont encore assez récentes, et peu de personnes s'y intéressent en tant qu'objet d'étude à part entière.

\end{point}

\begin{point}{La contribution proposée}

Mon apport est celui d'une démonstration de la correction de la seconde méthode de de Groote. Elle consiste principalement à réécrire une dérivation dans Lambek en dérivation équivalente ne contenant que des coupures. Pour arriver à cela, j'ai préalablement analysé les formes de dérivation initiales, dans le but de mettre à jour quels outils seraient nécessaires à la démonstration. Avec cela, j'ai pu déclarer clairement les définitions utiles (systèmes intermédiaires, relations, opérateurs,...), pour enfin réussir à établir la preuve.

\end{point}

\begin{point}{Les arguments en faveur de sa validité}

Ma démonstration montre la validité de la seconde méthode de de Groote. J'ai aussi écrit une ébauche d'algorithme implémentant cette solution, qui fonctionne sur des petits ensembles de français ou d'anglais. Son application à la main semble plus facile que la première méthode, ce qui joue en faveur de sa pertinence pragmatique. Par ailleurs, le résultat d'une transformation y est plus intéressant que par Pentus, puisque la grammaire produite est moins ambiguë (notamment plus petite) et plus logique linguistiquement. Ceci se vérifie aisément sur des petits exemples.

Cependant, la traduction n'est intéressante (voire possible) que pour la variante des grammaires de Lambek à laquelle je me suis restreint. De plus, l'application de l'algorithme requiert un certain type de grammaire de Lambek. Bien que presque tous les exemples linguistiques répondent à ce critère, cela n'est pas le cas d'un exemple quelconque. La solution n'est donc pas utilisable en pratique sur des entrées trop exotiques.

\end{point}

\begin{point}{Le bilan et les perspectives}

Ma contribution est la mise à disposition d'une méthode générique de traduction de la dérivation dans une grammaire de Lambek en celle dans une ACG dans les cas linguistiques. Une perspective serait d'étudier le cas des grammaires ne répondant pas au critère cité précédemment, et de fournir un procédé complémentaire. Une autre serait d'implémenter cet algorithme dans l'outil de l'équipe ACGtk.

\end{point}
\thispagestyle{empty}
\clearpage
\setcounter{page}{1}

\tableofcontents

\section{Contexte scientifique}

\subsection{Le Loria}

Le Loria (Laboratoire Lorrain de recherche en informatique et ses applications) est une unité mixte de recherche (UMR) regroupant des chercheurs en informatique affiliés à l'Inria, le CNRS ou l'Université de Lorraine. Comprenant 27 équipes de recherche (190 (enseignant-)chercheurs permanents et 100 doctorants, le tout de 50 nationalités différentes), ce laboratoire est très grand, prospère et dynamique, comme le montrent les 600 publications internationales depuis sa création, 9 lauréats ERC (European Research Council : bourses d'excellence scientifique) et les 14 start-ups présentes.

Les locaux sont situés à Vandœuvre-lès-Nancy, proches de la ligne de tramway, et sont divisés en 3 grands bâtiments connexes. On y trouve une cafétéria et un restaurant interne offrant une réduction intéressante pour les stagiaires d'équipes affiliées Inria. Le centre est de plus pourvu de toutes sortes de salles pour se détendre, par exemple la médiathèque.

\subsection{L'équipe SEMAGRAMME}

L'équipe SEMAGRAMME fait partie du département \href{http://www.loria.fr/fr/la-recherche/departements/4-traitement-des-langues-et-des-connaissances/}{4}, dont l'axe de recherche est la linguistique informatique et le traitement automatique des langues (TAL). Sa spécificité est le développement de modèles, méthodes et outils basés sur la logique, pour l'analyse sémantique d'énoncés et de discours en langue naturelle. Concrètement, elle propose des programmes d'analyse grammaticale (\href{https://gitlab.inria.fr/ACG/dev/ACGtk}{ACGtk}) ou sémantique, des modèles théoriques sémantiques du discours et de la conversation et même un jeu en ligne d'annotation de phrases (\href{https://zombilingo.org}{Zombilingo}). Ainsi les recherches visent autant à soutenir le travail des linguistes dans une approche théorique (modèles) que pratique (méthodes et outils).

Je travaillais dans le bureau des doctorants de l'équipe avec 2 autres stagiaires et 3 doctorants, juste à côté de ceux des permanents. L'ambiance était très agréable et je suis satisfait de ne pas avoir été mis dans une salle pour stagiaires. En effet, j'ai pu discuter avec les gens de mon équipe sur la linguistique et leur travaux respectifs. On s'entraidait même pour trouver des (contre-)exemples ou relire les productions des autres. Cela m'a vraiment donné un bon aperçu des conditions de travail d'une thèse, et plus généralement de la recherche. J'ai aussi beaucoup apprécié la grande autonomie qui m'a été accordée, de telle sorte que j'ai pu bien appréhender la mise en place et le développement des idées.

\subsection{La recherche sur les grammaires catégorielles}

Mon maître de stage Philippe de Groote est spécialisé dans les modèles logiques de modélisation de la langue. Il a notamment développé une sémantique dynamique en $\lambda$-calcul par passage de continuation, qui permet de formaliser les interactions entre les propositions d'un discours. Il étudie aussi les \textbf{grammaires catégorielles}, c'est-à-dire les systèmes créés pour le langage naturel, reposant sur le principe que les objets (souvent les mots) interagissent par composition, via des liaison fonction-argument. Elles sont à comparer notamment aux grammaires de dépendance. L'objectif d'une grammaire est en fait de résumer sous un petit nombre de règle l'ensemble de phrases qu'une langue accepte.

C'est aussi Philippe qui a créé les grammaires catégorielles abstraites (ACG), un modèle qui a pour objectif de généraliser les précédents modèles linguistiques. Afin de montrer cela, il propose à certains stagiaires et doctorants de l'aider à traduire les formalismes déjà existants dans les ACG, et ainsi argumenter pour la pertinence des ACG en linguistique informatique. C'est dans cette optique que se place mon travail, face aux grammaires de Lambek.

Les grammaires de Lambek ont été inventées en 1958 par Lambek pour tenter d'extraire la sémantique d'une phrase à partir de sa seule syntaxe. Elles font aujourd'hui référence en terme de grammaire catégorielle, et connaissent donc beaucoup de variantes (calcul syntactique à produit, grammaire non associative, avec types modaux,...). Cependant, leur expressivité est très restreinte et ne reste pertinente que pour certaines langues.

\section{Sujet de recherche}
\subsection{Grammaire de Lambek associative}

Une \textbf{grammaire (catégorielle) de Lambek associative} \cite{Lam58} est une grammaire catégorielle qui associe à chaque mot un (ou plusieurs) types, et accepte une phrase si et seulement si le mot des types se réécrit en un type distingué $s$. Elle mise donc tout sur la syntaxe. Plus formellement, étant donné un ensemble fini de symboles $\Pr$ appelés \textbf{types atomiques}, on défini un ensemble de \textbf{types} dit \textbf{orientés} par induction : 
\[ \Tp:\alpha,\beta,\gamma,\delta,...::=s,r,p,q,...\in\Pr~|~\beta/\alpha~|~\alpha\backslash\beta \]
Une grammaire de Lambek (LG) est alors un quadruplet $\m{G}=(\Pr,\m{T},\chi,s)$ où
\begin{itemize}
\item $\Pr$ est l'ensemble des types atomiques (aussi appelés types de base)
\item $\m{T}$ est l'ensemble des symboles terminaux, appelés \textbf{symboles lexicaux} ou lexèmes
\item $\chi:\m{T}\to\m{P}(\Tp)$ qui à chaque symbole lexical associe des types orientés
\item $s\in \Tp$ est un type distingué
\end{itemize}

On note $\Tp(\m{G})\triangleq\bigcup_{t\in\m{T}}\chi(t)$ l'ensemble des types d'un grammaire $\m{G}$, qui sera fixée pour le reste du rapport. Un exemple se trouve en table [\ref{fig:S_IE}].

La grammaire est munie d'un système de dérivation appelé calcul syntactique de Lambek. C'est par ce processus de dérivation qu'on détermine si une suite de mots est accepté ou non.

Un \textbf{système de dérivation} est un ensemble de règles formées d'une ou plusieurs prémisses (en haut de la barre) et d'une conclusion (en bas) sous une forme particulière (souvent une relation appelée séquent (noté $\vdash$ lu "thèse")  ou réécriture $\Rightarrow$). Une règle peut avoir une ou plusieurs conditions d'application, appelées conditions de bord. Si la règle n'a pas de prémisse, on l'appelle axiome. Effectuer une preuve (ou dérivation) dans un système de dérivation, c'est construire un arbre dont les nœuds sont des instances des règles (c'est-à-dire qui ont la forme d'une règle) et les feuilles des axiomes. La racine (traditionnellement en bas) est le séquent prouvé. La dérivation en figure [\ref{fig:der_G_0_IE}] en est un exemple pour le système de la table [\ref{fig:S_IE}].

Les grammaires de Lambek étant diverses, on ne s'intéresse qu'à une version particulière\footnote{Dans la littérature, cette version est dite associative et sans produit. Ici, on prend même une version alternative (équivalente) dite lexicalisée.} : le \textbf{système} $\m{S}_{IE}$. On considère des séquents de la forme $\Gamma\vdash\beta$ où $\Gamma$ (on utilise aussi $\Delta,\Phi,\Pi$ où $\Xi$) est un mot \underline{non vide} sur l'alphabet $\Tp\uplus\m{T}$ (avec une virgule ',' ou rien pour la composition de mots), $\beta\in\Tp$. Les règles sont détaillées dans la figure [\ref{fig:S_IE}].

\begin{figure}[ht]
\centering
\begin{minipage}{5cm}
\AxiomC{}
\RightLabel{(Ax.)}
\UnaryInfC{$\alpha\vdash\alpha$}
\DisplayProof
\newline\vspace{0.5cm}

\AxiomC{$\Gamma,\alpha\vdash\beta$}
\RightLabel{(/I)}
\UnaryInfC{$\Gamma\vdash\beta/\alpha$}
\DisplayProof\vspace{0.5cm}

\AxiomC{$\alpha,\Gamma\vdash\beta$}
\RightLabel{(\b I)}
\UnaryInfC{$\Gamma\vdash\alpha\backslash\beta$}
\DisplayProof
\end{minipage}
\begin{minipage}{5cm}
\AxiomC{$\alpha\in\chi(t)$}
\RightLabel{(Lex.)}
\UnaryInfC{$t\vdash\alpha$}
\DisplayProof
\newline\vspace{0.5cm}

\AxiomC{$\Gamma\vdash\beta/\alpha$}
\AxiomC{$\Delta\vdash\alpha$}
\RightLabel{(/E)}
\BinaryInfC{$\Gamma,\Delta\vdash\beta$}
\DisplayProof\vspace{0.5cm}

\AxiomC{$\Gamma\vdash\alpha\backslash\beta$}
\AxiomC{$\Delta\vdash\alpha$}
\RightLabel{(\b E)}
\BinaryInfC{$\Delta,\Gamma\vdash\beta$}
\DisplayProof
\end{minipage}
\caption{Règles du calcul de Lambek (déduction naturelle) sans produit lexicalisé $\m{S}_{IE}$\label{fig:S_IE}}
\end{figure}

Étant donné un système de dérivation $\m{S}_X$ sur ce genre de séquents, on notera $\Gamma\vdash_X\beta$ si $\Gamma\vdash\beta$ est \textbf{dérivable} dans $\m{S}_X$. Le \textbf{langage} (l'ensemble des mots acceptés) de $\m{G}$ est $\m{L}(\m{G})\triangleq\{\Gamma\in\m{T}^+~|~\Gamma\vdash_{IE}s\}$. Analyser grammaticalement (en anglais \textit{to parse}) le mot $\Gamma\in\m{T}^+$, c'est trouver une dérivation de $\Gamma\vdash s$. Un exemple de dérivation se situe en figure [\ref{fig:der_G_0_IE}].

La règle de \textbf{coupure} 
\AxiomC{$\Phi,\gamma,\Delta\vdash\beta$}
\AxiomC{$\Gamma\vdash\gamma$}
\RightLabel{(CUT)}
\BinaryInfC{$\Phi,\Gamma,\Delta\vdash\beta$}
\DisplayProof
pourrait être ajoutée à ce système, mais elle y est admissible (c'est-à-dire qu'elle peut être simulée avec les règles déjà présentes) (d'après Grentzen\footnote{On peut retrouver la preuve dans \cite{Lam58}}).

Pour plus de concision, on adopte une \textbf{écriture} plus \textbf{compacte} et plus générale : $/(\alpha,\beta)\triangleq\beta/\alpha$, $\backslash(\alpha,\beta)\triangleq\alpha\backslash\beta$, $f_/(\Gamma,\Delta)\triangleq\Gamma,\Delta$ et $f_{\backslash}(\Gamma,\Delta)\triangleq\Delta,\Gamma$. On prend de manière systématique la lettre $c$ pour $c\in\{/,\backslash\}$, et on note $\tilde{c}$ la barre complémentaire de $c$. On peut ainsi réduire notre système à seulement deux règles (plus les deux axiomes) en figure [\ref{fig:S_IE_comp}] avec une nouvelle version de la coupure.

\begin{figure}[ht]
\centering
\begin{minipage}{4cm}
\AxiomC{$f_c(\Gamma,\alpha)\vdash\beta$}
\RightLabel{($c$ I)}
\UnaryInfC{$\Gamma\vdash c(\alpha,\beta)$}
\DisplayProof
\end{minipage}
\begin{minipage}{5cm}
\AxiomC{$\Gamma\vdash c(\alpha,\beta)$}
\AxiomC{$\Delta\vdash\alpha$}
\RightLabel{($c$ E)}
\BinaryInfC{$f_c(\Gamma,\Delta)\vdash\beta$}
\DisplayProof
\end{minipage}
\begin{minipage}{6cm}
\AxiomC{$f_c(f_c(\Phi,\gamma),\Delta)\vdash\beta$}
\AxiomC{$\Gamma\vdash\gamma$}
\RightLabel{(CUT)}
\BinaryInfC{$f_c(f_c(\Phi,\Gamma),\Delta)\vdash\beta$}
\DisplayProof
\end{minipage}
\caption{Règles non axiomatiques sous forme compact de $\m{S}_{IE}$\label{fig:S_IE_comp}}
\end{figure}

Insistons sur le fait que l'écriture $f_c(\Gamma,\Delta)$ cache en réalité un mot, sur lesquels la composition est associative, d'où les règles d'associativité et autre en figure [\ref{fig:assoc}].

\begin{figure}[ht]
\centering
$\begin{array}{ccc} f_c(\Gamma,\Delta)=f_{\tilde{c}}(\Delta,\Gamma) & f_c(\Gamma,\varepsilon)=\Gamma & f_c(f_c(\Phi,\Gamma),\Delta)=f_c(\Phi,f_c(\Gamma,\Delta)) \end{array}$
\caption{Formules d'associativité et autres de l'écriture compact\label{fig:assoc}}
\end{figure}

Enfin, on supposera jusqu'à la fin que le type distingué $s$ est atomique, ce à quoi on peut toujours se ramener en ajoutant un lexème '.' (point final) de type $s\backslash s'$, où $s'$ est un symbole atomique frais rajouté.

\subsection{\texorpdfstring{$\lambda$}{lambda}-calcul et typage}

Les ACG basent leur modèle sur le $\lambda$-calcul typé. Cet outil informatique est de plus justement pertinent pour étudier certains systèmes de dérivation, comme nous allons le voir.

Rappelons donc rapidement le \textbf{$\lambda$-calcul avec constantes}. Donnons-nous un ensemble infini dénombrable $\m{X}$ de variables et un ensemble de constantes C. On appelle $\lambda$-terme un objet construit sur
\begin{equation} \label{eq:lambda}
\Lambda(C):u,v,w...::= x,y,z...\in\m{X}~|~t,...\in C~|~\uabs{x}{u}~|~\app{u}{v}
\end{equation}
Intuitivement $\uabs{x}{u}$ est une fonction qui à un argument $x$ associe $u$, et $\app{u}{v}$ l'application de $u$ à $v$. On adopte le parenthésage conventionnel $\app{\app{u}{v}}{w}\triangleq\app{(\app{u}{v})}{w}$. Exemple avec une constante \textsc{a} : $v_0\triangleq\uabs{x,y}{\app{\app{\uabs{z}{\app{\textsc{a}}{z}}}{x}}{y}}$\footnote{$\uabs{x,y}{u}$ est un diminutif pour $\uabs{x}{\uabs{y}{u}}$}.

On munit les $\lambda$-termes d'une notion de $\beta$-réduction ($\to_{\beta}$) et de $\eta$-réduction ($\to_{\eta}$) (et leur inverse $\beta$- et $\eta$-expansion) décrites par les règles en figure [\ref{fig:beta_eta}].
\begin{figure}[ht]
\centering

Axiomes\vspace{0.5cm}

\begin{minipage}{5cm}
\AxiomC{}
\UnaryInfC{$\app{\uabs{x}{u}}{v}\to_{\beta}u[x:=v]$}
\DisplayProof
\end{minipage}
\begin{minipage}{5cm}
\AxiomC{$x$ non libre dans $u$}
\UnaryInfC{$\uabs{x}{\app{u}{x}}\to_{\eta}u$}
\DisplayProof
\end{minipage}\vspace{0.5cm}

Passage au contexte ($X\in\{\beta,\eta\}$)\vspace{0.5cm}

\begin{minipage}{4cm}
\AxiomC{$u\to_Xu'$}
\UnaryInfC{$\uabs{x}{u}\to_X\uabs{x}{u'}$}
\DisplayProof
\end{minipage}
\begin{minipage}{4cm}
\AxiomC{$u\to_Xu'$}
\UnaryInfC{$\app{u}{v}\to_X\app{u'}{v}$}
\DisplayProof
\end{minipage}
\begin{minipage}{4cm}
\AxiomC{$v\to_Xv'$}
\UnaryInfC{$\app{u}{v}\to_X\app{u}{v'}$}
\DisplayProof
\end{minipage}
\caption{Règles de $\beta$- et $\eta$-réduction\label{fig:beta_eta}}
\end{figure}
,avec les bonnes notions de variables libres et liées, d'$\alpha$-équivalence, $\alpha$-renommage et substitution $u[x:=v]$ que je ne ré-explicite pas ici, mais qui sont disponibles dans \cite{Bar84}. Intuitivement, ces réductions représente une étape de calcul, mais on se contentera de regarder à $\beta$- et $\eta$-équivalence près\footnote{C'est-à-dire considérer les classes de la relation rendu réflexive, symétrique et transitive}. Exemple : $v_0\to_{\eta}\uabs{x}{\app{\uabs{z}{\app{\textsc{a}}{z}}}{x}}\to_{\beta}\uabs{x}{\app{\textsc{a}}{x}}\to_{\eta}\textsc{a}$.

On peut donner à certains $\lambda$-termes un \textbf{type non orienté} formé par induction à partir d'un ensemble de types de base $\m{B}$ :
\begin{equation} \label{eq:Tp}
\Tp_{no}(\m{B}):\alpha,\beta,\gamma,\delta,...::= s,r,p,q,...\in\m{B}~|~\alpha\to\beta
\end{equation}
avec les règles $\m{S}_{no}$ en figure [\ref{fig:typage_no}] sur des séquents de la forme $\Gamma\vdash\beta$ avec $\Gamma$ un ensemble de couples $(\m{X}:\Tp_{no})$ et $\beta\in\Tp_{no}$ où pour $x$ donné, il existe au plus un couple où $x$ intervient. Ici la virgule symbolise l'union d'ensembles. Il faut de plus se donner une fonction $\chi_{no}:\m{T}\to\m{P}(\Tp_{no})$ qui associe un ou plusieurs types aux constantes. L'assertion $\Gamma\vdash\beta$ (le mot $\Gamma$ a le type $\beta$) se prouve donc (s'il est prouvable) par une dérivation. Ainsi, on peut dire qu'un $\lambda$-terme $u$ typable (noté $\vdash_{no}u:\beta$) représente la dérivation de son type $\beta$. Cela nous permettra de mieux visualiser une dérivation, grâce à seul son $\lambda$-terme. Par exemple, si $r\to q\in\chi(\textsc{a})$, $\uabs{x}{\app{\textsc{a}}{x}}$ est dérivable dans $\m{S}_{no}$ :

\AxiomC{}
\RightLabel{(Lex.$_{no}$)}
\UnaryInfC{$x:r\vdash\textsc{a}:r\to q$}
\AxiomC{}
\RightLabel{(Ax.$_{no}$)}
\UnaryInfC{$x:r\vdash x:r$}
\RightLabel{($\to$E$_{no}$)}
\BinaryInfC{$x:r\vdash \app{\textsc{a}}{x}:q$}
\RightLabel{($\to$I$_{no}$)}
\UnaryInfC{$\vdash\uabs{x}{\app{\textsc{a}}{x}}:r\to q$}
\DisplayProof

\begin{figure}[ht]
\centering
\begin{minipage}{4cm}
\AxiomC{}
\RightLabel{(Ax.$_{no}$)}
\UnaryInfC{$\Gamma,x:\alpha\vdash x:\alpha$}
\DisplayProof
\end{minipage}
\begin{minipage}{4cm}
\AxiomC{$\alpha\in\chi_{no}(t)$}
\RightLabel{(Lex.$_{no}$)}
\UnaryInfC{$\Gamma\vdash t:\alpha$}
\DisplayProof
\end{minipage}\vspace{0.5cm}

\begin{minipage}{5cm}
\AxiomC{$\Gamma,x:\alpha\vdash u:\beta$}
\RightLabel{($\to$I$_{no}$)}
\UnaryInfC{$\Gamma\vdash \uabs{x}{u}:\alpha\to\beta$}
\DisplayProof
\end{minipage}
\begin{minipage}{6cm}
\AxiomC{$\Gamma\vdash u:\alpha\to\beta$}
\AxiomC{$\Gamma\vdash v:\alpha$}
\RightLabel{($\to$E$_{no}$)}
\BinaryInfC{$\Gamma\vdash \app{u}{v}:\beta$}
\DisplayProof
\end{minipage}
\caption{Règles de typage simple (non orienté) $\m{S}_{no}$ du $\lambda$-calcul avec constantes\label{fig:typage_no}}
\end{figure}

À tout type $\alpha$ on peut associer un entier $\ord(\alpha)$ appelé ordre qui quantifie à quel point les $\to$ sont "imbriquées" dans $\alpha$, par induction :
\[ \begin{array}{cl}
\ord(r)= 1 & \text{si $r\in\m{B}$} \\
\ord(\alpha\to\beta)=\max\{1+\ord(\alpha),\ord(\beta)\} &
\end{array} \]
et de même pour les types orientés. C'est un indicateur de la "complexité" d'un type.

On utilisera par la suite un représentant de la classe de $\beta\eta$-équivalence (donc une dérivation particulière parmi toutes les possible pour une même séquent) qui a des propriétés intéressantes. Cela découle des propriétés du typage. Commençons quelques définitions :

Un $\lambda$-terme $v$ est dit $\beta$-normal si pour tout $v$, $u\not\to_{\beta} v$. On appelle contexte $u[]$ tout $\lambda$-terme $u$ ayant exactement une seule occurrence libre d'une variable distinguée $\square$ : le trou, et on écrit $u[v]$ pour $u[\square:=v]$. On dit que $u$ est en forme $\eta$-longue si pour toute décomposition $u=u'[v]$ avec $v$ de type $\alpha\to\beta$, on a $v$ qui vérifie l'une des deux conditions suivantes:
\begin{itemize}
\item $u'[]$ est de la forme $u''[\app{\square}{v'}]$
\item $v$ est de la forme $\uabs{x}{v'}$
\end{itemize}
autrement dit, soit $v$ est appliqué à un argument soit $v$ est un argument et commence par abstraire la variable de son argument (du type $\alpha$). On a le résultat suivant : tout $\lambda$-terme typable (c'est-à-dire tel que $\vdash u:\beta$ est dérivable pour un certain $\beta$) admet (est $\beta\eta$-équivalent à) une unique \textbf{forme $\beta$-normale $\eta$-longue} notée $u\!\downarrow_{\beta\eta}$. Par exemple, pour $\textsc{a}:r\to q$, $v_0\!\downarrow_{\beta\eta}=\uabs{x}{\app{\textsc{a}}{x}}$.

Enfin, on ne considérera ici que des $\lambda$-termes linéaires, c'est-à-dire où chaque variable apparaît au plus une fois liée. On note alors habituellement le type flèche $\alpha\multimap\beta$, mais  on s'en passera dans ce rapport puisqu'il n'y a pas ambiguïté. Les règles de typage y sont aussi légèrement différentes, mais ces particularités sont présentes dans le typage que nous allons introduire maintenant.

\subsubsection{\texorpdfstring{$\lambda$}{lambda}-terme d'une dérivation de Lambek}

Remarquons que le type orienté $c(\alpha,\beta)$ se comporte comme un type flèche non orienté $\alpha\to\beta$, typant donc une fonction d'argument de type $\alpha$ et de type de retour $\beta$. On peut même décomposer tout type orienté $\alpha$ de manière unique en $\alpha=c_1(\alpha_1,c_2(\alpha_2,...c_n(\alpha_n,r)...))$ d'arguments $\alpha_1,...\alpha_n$ et de type de retour atomique $r$. Fort de cette analogie, on peut associer à toute dérivation dans $\m{S}_{IE}$ un $\lambda$-terme sur les constantes $\m{T}$ et types atomiques $\Pr$, d'après le typage correspondant présent en figure [\ref{fig:typage}]. Les séquents sont alors de la forme $\Gamma\vdash u:\beta$ avec $\Gamma$ un mot \underline{non vide} sur l'alphabet $(\m{X}:\Tp)\uplus\m{T}$ (où ':' est un symbole frais), $u$ un $\lambda$-terme et $\beta\in\Tp$.

\begin{figure}[ht]
\centering
\begin{minipage}{5cm}
\AxiomC{}
\RightLabel{(Ax.)}
\UnaryInfC{$x:\alpha\vdash x:\alpha$}
\DisplayProof
\vspace{0.5cm}

\AxiomC{$f_c(\Gamma,x:\alpha)\vdash u:\beta$}
\RightLabel{($c$ I)}
\UnaryInfC{$\Gamma\vdash \uabs{x}{u}:c(\alpha,\beta)$}
\DisplayProof
\end{minipage}
\begin{minipage}{5cm}
\AxiomC{$\alpha\in\chi(t)$}
\RightLabel{(Lex.)}
\UnaryInfC{$t\vdash t:\alpha$}
\DisplayProof
\vspace{0.5cm}

\AxiomC{$\Gamma\vdash u:c(\alpha,\beta)$}
\AxiomC{$\Delta\vdash v:\alpha$}
\RightLabel{($c$ E)}
\BinaryInfC{$f_c(\Gamma,\Delta)\vdash\app{u}{v}:\beta$}
\DisplayProof
\end{minipage}\vspace{0.5cm}

\AxiomC{$f_c(f_c(\Phi,y:\gamma),\Delta)\vdash u:\beta$}
\AxiomC{$\Gamma\vdash v:\gamma$}
\RightLabel{(CUT)}
\BinaryInfC{$f_c(f_c(\Phi,\Gamma),\Delta)\vdash u[y:=v]:\beta$}
\DisplayProof
\caption{Typage d'un $\lambda$-terme dans $\m{S}_{IE}$\label{fig:typage}}
\end{figure}

Comme il s'agit d'un sous-typage du typage simple non orienté (c'est-à-dire que tout terme typable dans $\m{S}_{IE}$ l'est dans $\m{S}_{no}$), on conserve les propriétés énoncées précédemment.

\subsection{Exemple}

Prenons la grammaire $\m{G}_0$ définie sur $\Pr\triangleq\{s,n,np\}$ dans la table [\ref{tab:G_0}] (où ici $\chi:\m{T}\to\Tp$). Intuitivement, $s$ est le type de la phrase (\textit{sentence}), $n$ représente un nom et $np$ un groupe (syntagme) nominale (\textit{noun phrase}). Un verbe transitif a le type $(np\backslash s)/np$, c'est-à-dire prend à droite un objet (C.O.D) puis à gauche un sujet. On a alors la phrase \textsc{le chat que Pierre voit dort} qui appartient au langage de $\m{G}_0$, et sa dérivation ($\beta$-normale $\eta$-longue) dans $\m{S}_{IE}$ en figure [\ref{fig:der_G_0_IE}] (seules les initiales ont été laissées pour faire tenir la dérivation sur la largeur).

\begin{table}[ht]
\centering

\fbox{
\begin{tabular}{rlrlrl}
\textsc{le} : & $np/n$ & \textsc{chat} : & $n$ & \textsc{dort} : & $np\backslash s$ \\
\textsc{que} : & $(n\backslash n)/(s/np)$ & \textsc{Pierre} : & $np$ & \textsc{voit} : & $(np\backslash s)/np$
\end{tabular}
}
\caption{Exemple 0 : Grammaire de Lambek $\m{G}_0$\label{tab:G_0}}
\end{table}

\begin{figure}[ht]
\hspace{-1cm}
\footnotesize
\AxiomC{$\textsc{d}\vdash \textsc{d}:np\backslash s$}
	\AxiomC{$\textsc{l}\vdash \textsc{l}:np/ n$}
		\AxiomC{$\textsc{q}\vdash \textsc{q}:(n\backslash n)/(s/np)$}
			\AxiomC{$\textsc{v}\vdash \textsc{v}:(np\backslash s)/np$}
			\AxiomC{$x:np\vdash x:np$}
			\RightLabel{(/E)}
			\BinaryInfC{$\textsc{v},x:np\vdash \app{\textsc{v}}{x}:np\backslash s$}
			\AxiomC{$\textsc{P}\vdash\textsc{P}:np$}
			\RightLabel{(\b E)}
			\BinaryInfC{$\textsc{P,v},x:np\vdash \app{\app{\textsc{v}}{x}}{\textsc{P}}:s$}
		\RightLabel{(/I)}
		\UnaryInfC{$\textsc{P,v}\vdash \uabs{x}{\app{\app{\textsc{v}}{x}}{\textsc{P}}}:s/np$}
		\RightLabel{(/E)}
		\BinaryInfC{$\textsc{q,P,v}\vdash \app{\textsc{q}}{\uabs{x}{\app{\app{\textsc{v}}{x}}{\textsc{P}}}}:n\backslash n$}
		\AxiomC{$\textsc{c}\vdash\textsc{c}:n$}
		\RightLabel{(\b E)}
		\BinaryInfC{$\textsc{c,q,P,v}\vdash \app{\app{\textsc{q}}{\uabs{x}{\app{\app{\textsc{v}}{x}}{\textsc{P}}}}}{\textsc{c}}:n$}
	\RightLabel{(/E)}
	\BinaryInfC{$\textsc{l,c,q,P,v}\vdash \app{\textsc{l}}{\app{\app{\textsc{q}}{\uabs{x}{\app{\app{\textsc{v}}{x}}{\textsc{P}}}}}{\textsc{c}}}:np$}
\RightLabel{(\b E)}
\BinaryInfC{$\textsc{l,c,q,P,v,d}\vdash \app{\textsc{d}}{\app{\textsc{l}}{\app{\app{\textsc{q}}{\uabs{x}{\app{\app{\textsc{v}}{x}}{\textsc{P}}}}}{\textsc{c}}}}:s$}
\DisplayProof
\caption{Exemple 0 : Dérivation ($\beta$-normale $\eta$-longue) de \textsc{le chat que Pierre voit dort} de $\m{G}_0$ dans $\m{S}_{IE}$\label{fig:der_G_0_IE}}
\end{figure}

On pourrait étendre cette grammaire avec des adjectifs, comme \textsc{beau}$:n/n$ et des adverbes, comme \textsc{très}$:(n/n)/(n/n)$, pour pouvoir dériver $\textsc{le,très,beau,chat,dort}\vdash\app{\textsc{d}}{\app{\textsc{l}}{\app{\textsc{t}}{\app{\textsc{b}}{\textsc{c}}}}}:s$. Mais si on acceptait le mot vide à gauche du séquent, ce dernier pourrait avoir le type $\varepsilon\vdash\uabs{x}{x}:n/n$, et donc on pourrait dériver \textsc{le très chat dort}, ce qui n'est pas grammaticalement correct. De plus, cela pose de gros problèmes d'analyse grammaticale. C'est pourquoi on interdit le mot vide à gauche du séquent.

\subsection{Grammaire catégorielle abstraite}

Pour définir les \textbf{grammaires catégorielles abstraites} (ACG) \cite{dG01}, revenons sur les types non orientés\footnote{De même, on demande et travaillera avec des types linéaires, mais on garde la notation $\to$ au lieu de $\multimap$} $\Tp_{no}(\m{B})$ à partir d'un ensemble fini de types atomiques $\m{B}$, d'après l'équation (\ref{eq:Tp}). Une \textbf{signature d'ordre supérieur} est un triplet $\Sigma=(\m{B},C,\tau)$ où
\begin{itemize}
\item $\m{B}$ est un ensemble fini de types atomiques (aussi dit de base)
\item $C$ est un ensemble fini de constantes
\item $\tau:C\to\Tp_{no}(\m{B})$ assigne un type à chaque constante
\end{itemize}

Une signature d'ordre supérieur (ou vocabulaire) représente un ensemble de mots (les constantes) et la manière dont ils peuvent se composer entre eux (les types), comme dans l'exemple en table [\ref{tab:voc_G_0}] où on peut dériver le $\lambda$-terme $\app{\textsc{dort}}{\app{\textsc{le}}{\textsc{chat}}}\vdash s$.

\begin{table}[ht]
\centering
\fbox{
\begin{tabular}{rlrlrl}
\textsc{le}: & $n\to np$ & \textsc{chat}: & $n$ & \textsc{dort}: & $np\to s$
\end{tabular}
}
\caption{Lexique exemple (de la forme \textsc{constant}:~~$type$) sur quelques termes de l'exemple 0\label{tab:voc_G_0}}
\end{table}

Un \textbf{lexique} de $\Sigma_1$ vers $\Sigma_2$ est un morphisme de signature d'ordre supérieur, autrement dit un couple $\mathfrak{L}=(F,G)$ où
\begin{itemize}
\item $F:\m{B}_1\to\Tp_{no}(\m{B}_2)$ (on note $\hat{F}:\Tp_{no}(\m{B}_1)\to\Tp_{no}(\m{B}_2)$ l'homomorphisme induit)
\item $G:c_1\to\Lambda(C_2)$ (on note $\hat{G}:\Lambda(C_1)\to\Lambda(C_2)$ l'homomorphisme induit)
\item pour tout $c\in C_1$, on a $\vdash_{no}G(c):F(\tau_1(c))$
\end{itemize}
avec $\Lambda(C_i)$ défini par l'équation (\ref{eq:lambda}). On peut noter directement (abusivement) $\mathfrak{L}$ au lieu de $\hat{F}$ ou $\hat{G}$. La section \ref{sec:ex_lex} montre un exemple de lexique.

On peut maintenant définir une grammaire catégorielle abstraite par $\mathfrak{G}=(\Sigma_1,\Sigma_2,\mathfrak{L},s)$ où
\begin{itemize}
\item $\Sigma_1$ (resp. $\Sigma_2$) est une signature d'ordre supérieur appelé le \textbf{vocabulaire abstrait} (resp. \textbf{objet})
\item $\mathfrak{L}:\Sigma_1\to\Sigma_2$ est un lexique de $\Sigma_1$ vers $\Sigma_2$.
\item $s\in\Tp_{no}(\m{B}_1)$ est un type distingué
\end{itemize}

On définit aussi le \textbf{langage abstrait} par
\[ \mathfrak{A}(\mathfrak{G})\triangleq\{u\in\Lambda(C_1)~|~\vdash_{no}u:s\} \]
et celui auquel on s'intéresse vraiment, le \textbf{langage objet} $\mathfrak{O}(\mathfrak{G})\triangleq\mathfrak{L}(\mathfrak{A}(\mathfrak{G}))$. Analyser grammaticalement un $\lambda$-terme objet $v$ revient à trouver un antécédent dérivable $u$ de $v$ par $\mathfrak{L}$. Autrement dit, il suffit d'inverser le morphisme $\mathfrak{L}$, ce qui est possible en temps polynomial dès lors que le vocabulaire abstrait $\Sigma_1$ est d'ordre inférieur à 2, où
\[ \ord(\Sigma_1)\triangleq\max_{c\in C_1} \ord(\tau_1(c)) \]

\section{Traductions naïve et via Pentus}

\subsection{Lexique de rendu}
\label{sec:ex_lex}

En poussant jusqu'au bout la remarque pour l'introduction du $\lambda$-terme d'une dérivation de Lambek, une traduction facile semble de transformer chaque type orienté en son homologue non orienté par $\tau_0$ :
\[ \begin{array}{cl}
\tau_0(r)=r & \text{si $r\in\Pr$} \\
\tau_0(c(\alpha,\beta))=\tau_0(\alpha)\to\tau_0(\beta) &
\end{array} \]
pour alors construire le vocabulaire (abstrait) $\Sigma_1\triangleq(\Pr,C_1,\tau_1)$, où $C_1\triangleq\{(t,\alpha)~|~t\in\m{T},\alpha\in\chi(t)\}$, et $\tau_1:(t,\alpha)\mapsto\tau_0(\alpha)$.

Il s'agit alors de retourner une chaîne de caractère correspondant au mot dont le $\lambda$-terme (abstrait) est celui de la dérivation du séquent dans $\m{S}_{IE}$. Pour ce faire, définissons une chaîne de caractères comme un type $\sigma\triangleq*\to*$, où '$*$' est un type atomique. On peut alors prendre $\Sigma_{\sigma}\triangleq(\{*\},C_1,\tau_2)$, avec $(t,\alpha)$ interprété comme la chaîne de caractères $t$, et $\tau_2(t,\alpha)=\tau_2'(\alpha)$ où $\tau_2'$ est le morphisme (de type orienté vers non orienté) tel que $\tau_2'(r)=\sigma$ si $r\in\Tp$.

Le lexique de l'ACG naïf est alors le lexique de rendu (yield) $\mathfrak{Y}\triangleq(r\mapsto\sigma,G_1)$, avec $G_1:(t,\alpha)\mapsto\mathbb{E}_{\alpha}t$ où l'opérateur $\mathbb{E}_{\alpha}$ et son conjoint $\mathbb{P}_{\alpha}$ sont trop techniques pour être explicités ici avec leurs propriétés, mais le sont dans \cite{dG16}. Ils utilisent l'opérateur de concaténation de chaînes (noté de manière infixe) $+\triangleq\uabs{f,g}{\uabs{z}{\app{f}{\app{g}{z}}}}$, et $\varepsilon\triangleq\uabs{x}{x}$ le mot vide. Contentons-nous de voir l'effet de $G_1$ sur notre grammaire exemple $\m{G}_0$ en table \ref{tab:G_0_Y}.

\begin{table}[ht]
\centering
\fbox{
\begin{tabular}{rlrlrl}
\textsc{le} : & $\uabs{x}{\mathit{le}+x}$ & \textsc{chat} : & $\mathit{chat}$ & \textsc{dort} : & $\uabs{x}{x+\mathit{dort}}$ \\
\textsc{que} : & $\uabs{y,x}{x+\mathit{que}+(\app{y}{\varepsilon})}$ & \textsc{Pierre} : & $\mathit{Pierre}$ & \textsc{voit} : & $\uabs{y,x}{x+\mathit{voit}+y}$
\end{tabular}
}
\caption{Exemple 0 : Table de $G_1$ pour la grammaire $\m{G}_0$\label{tab:G_0_Y}}
\end{table}

\begin{wrapfigure}[14]{R}{4cm}
\vspace{-0.5cm}
\begin{tikzpicture}[scale=1]
\node[draw,shape=ellipse] (A)at(0,4) {$\Sigma_0$};
\node[draw,shape=ellipse] (B)at(0,2) {$\Sigma_1$};
\node[draw,shape=ellipse] (C)at(0,0) {$\Sigma_{\sigma}$};
\draw[->,>=latex] (A) to[right] node {$\mathfrak{L}_{DER}$} (B) to[right] node {$\mathfrak{Y}$} (C);
\draw[->,>=latex] (A) to[left,bend right] node {$\mathfrak{L}_A$} (C);
\end{tikzpicture}
\caption{Schéma de composition des lexiques envisagés\label{fig:acg}}
\end{wrapfigure}
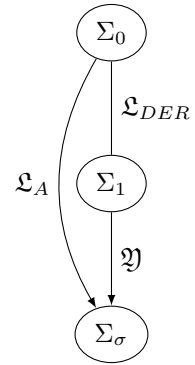

Cependant, l'ACG ainsi créé ne fonctionne pas : il surgénère. En effet, on n'implémente pas l'orientation des arguments, ce qui fait défaut dans leur appel en argument. Par exemple, dans le type non orienté \textsc{que}:$(np\to s)\to n\to n$, on ne sait pas si l'argument de type $np\to s$ prend son argument à droite ou à gauche.  Ainsi, on peut dériver dans cet ACG $\mathit{le+chat+que+voit+Pierre+dort}$, où il faut comprendre $\mathit{Pierre}$ comme l'objet du verbe $\mathit{voit}$\footnote{En fait, si la phrase sonne tout de même correcte ici, c'est dû à la particularité de la langue française à pouvoir inverser le verbe et le sujet dans ce cas. Si on avait utilisé $\mathit{je}$ au lieu de $\mathit{Pierre}$, si on avait fait une relative avec $\mathit{qui}$ et sans objet ou si on avait fait l'exemple en anglais, la phrase aurait été incorrecte.}. Or cette phrase n'est pas dérivable dans Lambek. De plus la traduction ne prend pas en compte l'ordre des arguments, comme le montre l'exemple en annexe \ref{an:naif}.

\subsection{Lexique de dérivations et limites}

Pour palier à cela, l'idée est d'ajouter une signature d'ordre supérieur $\Sigma_0$ avant $\Sigma_1$ qui doit rendre compte des dérivations possibles. On ne rentrera pas ici dans les détails techniques proposés par de Groote \cite{dG16}, mais il utilise notamment le résultat de Pentus \cite{Pen97} : toute grammaire de Lambek peut être transformée en une grammaire algébrique de même langage. Il s'agit même d'une équivalence (dite faible). Réintroduisons rapidement cette notion.

Une \textbf{grammaire algébrique} $\m{G}_A=(N,A,P,s)$ comprend
\begin{itemize}
\item $N$ un alphabet fini de symboles non terminaux $(K,L,...)$
\item $A$ un alphabet fini de symboles terminaux $(a,b,...)$, et on prend $N\cap A=\emptyset$
\item $P\in\m{P}(N\times (N\cup A)^*)$ un ensemble de productions notées $K\to w$
\item $s\in N$ un symbole distingué
\end{itemize}
et est munie d'un système de réécriture de règle le passage au contexte de $P$ :
\AxiomC{$K\to w\ \in P$}
\UnaryInfC{$w_1Kw_2\Rightarrow w_1ww_2$}
\DisplayProof
. On note $\Rightarrow^*$ la clôture réflexive transitive de $\Rightarrow$. Le langage de la grammaire algébrique est alors $\m{L}(\m{G}_A)\triangleq\{w\in A^*~|~s\Rightarrow^*w\}$. On dit enfin qu'un symbole non terminal K (de même pour les productions $K\to w$) est \textbf{accessible} s'il existe une dérivation $s\Rightarrow^*w_1Kw_2$.

Avec les productions de la grammaire algébrique de Pentus, de Groote construit une ACG par composition $\mathfrak{G}_A\triangleq(\Sigma_0,\Sigma_{\sigma},\mathfrak{Y}\circ\mathfrak{L}_{DER},s)$, après avoir opéré quelques autres transformations (lexicalisation, réduction de l'ordre du lexique). Cela lui assure, notamment grâce au résultat de Kanazawa et Salvati \cite{KanSal12}, que les $\lambda$-termes de $\mathfrak{G}_A$ correspondent aux dérivations initiales de Lambek (dans $\m{S}_{IE}$). 
Par conséquent, cette méthode constitue une première réponse à mon sujet.

Néanmoins, la construction précédente est insatisfaisante. En effet, les productions de Pentus sont très nombreuses et redondantes, et constituent une grammaire algébrique très ambiguë (c'est-à-dire qu'il existe beaucoup de dérivations possibles d'un même mot). Particulièrement, un grand nombre de productions sont non pertinentes d'un point de vue linguistique (exemple dans le dernier paragraphe de la section \ref{sec:ex_non_pert}).

Malgré tout, l'idée est bonne. Il peut donc être intéressant de passer de même par une grammaire algébrique, mais plutôt par une méthode plus algorithmique qui permettrait de construire pas à pas les productions utiles.

\section{Transformation des règles en coupure}

\subsection{Suppression des éliminations}

Reprenons l'idée de Pentus de transformer une grammaire de Lambek en grammaire algébrique. Pour cela, il remarque que la règle de coupure correspond exactement à la réécriture, dans les grammaires algébriques, sur des productions de la forme $\beta\to\Gamma$ pour un axiome $\Gamma\vdash\beta$. Mais contrairement à lui, adoptons une méthode plus constructive qui puisse donner plus de "sens linguistique" aux productions créées. Il s'agit donc de transformer les règles d'élimination et d'introduction en coupures, en transformant les axiomes (et en en ajoutant).

Dans un premier temps, supprimons les éliminations. L'idée est encore d'aplatir les types, mais dans le séquent directement. Formellement, pour un lexème $t$ de type décomposée de manière unique en $\alpha=c_1(\alpha_1,c_2(\alpha_2,...c_n(\alpha_n,r)...))$ où $r\in\Pr$, on appelle mot aplati le mot à trou $\pi(\alpha) = f_{c_n}(...f_{c_2}(f_{c_1}(\square,z_1:\alpha_1),z_2:\alpha_2),...z_n:\alpha_n)$, et son type de renvoi $\rho(\alpha) = r$. Cela forme des nouveaux axiomes (appelés \textbf{axiomes propres}) sur lesquels on peut couper, de la même manière qu'on élimine $c_i$.

Malheureusement, cela ne suffit pas à engendrer le langage du système initial à partir de l'ordre 4. Il faut rajouter des axiomes propres non lexicaux correspondants aux arguments des fonctions prises en argument. Plus clairement, définissons le \textbf{signe d'une occurrence d'une sous-formule} d'un type $\alpha$ en figure [\ref{fig:subset}].

\begin{figure}[ht]
\[\begin{array}{rcl}
 & \alpha\subset^+\alpha  & \\
\text{si }c(\gamma,\beta)\subset^+\alpha & \text{alors } \beta\subset^+\alpha & \text{et } \gamma\subset^-\alpha \\
\text{si }c(\gamma,\beta)\subset^-\alpha & \text{alors } \beta\subset^-\alpha & \text{et } \gamma\subset^+\alpha \\
\end{array}\]
\caption{Définition du signe d'une occurrence d'une sous formule\label{fig:subset}}
\end{figure}

Remarquons que $\subset^+$ est une relation d'ordre sur $\Tp$. On note $\Tp(\m{G})^+$ (resp. $\Tp(\m{G})^-$) l'ensemble des sous-types positifs (resp. négatifs) de la grammaire et
\[ \Tp(\m{G})^{+,\neq}\triangleq\{\alpha\in\Tp(\m{G})~|~\exists t\in\m{T},\exists \beta\in\chi(t),\alpha\subset^+\beta\wedge\alpha\neq\beta\} \]
les sous-types strictement positifs. L'ensemble des axiomes propres est alors :
\[ \m{A}_0\triangleq\{\pi(\alpha)[t]\vdash \app{t}{z_1...z_n}: \rho(\alpha)~|~t\in\m{T},\alpha\in\chi(t)\} \cup \{\pi(\alpha)[x:\alpha]\vdash \app{x}{z_1...z_n}:\rho(\alpha)~|~\alpha\in\Tp(\m{G})^{+,\neq}\} \]
On peut alors monter notre système intermédiaire $\m{S}_{IC}$ sans élimination (figure [\ref{fig:S_IC}]).

\begin{figure}[ht]
\centering
\begin{minipage}{6cm}
\AxiomC{$\alpha\in\chi(t)$}
\RightLabel{(pAx.$_{E1}$)}
\UnaryInfC{$\pi(\alpha)[t]\vdash \app{t}{z_1...z_n}:\rho(\alpha)$}
\DisplayProof\vspace{0.5cm}

\AxiomC{$f_c(\Gamma,\alpha)\vdash\beta$}
\RightLabel{($c$ I)}
\UnaryInfC{$\Gamma\vdash c(\alpha,\beta)$}
\DisplayProof
\end{minipage}
\begin{minipage}{6cm}
\AxiomC{$\alpha\in\Tp(\m{G})^{+,\neq}$}
\RightLabel{(pAx.$_{E2}$)}
\UnaryInfC{$\pi(\alpha)[x:\alpha]\vdash \app{x}{z_1...z_n}:\rho(\alpha)$}
\DisplayProof\vspace{0.5cm}

\AxiomC{$f_c(f_c(\Phi,y:\gamma),\Delta)\vdash u:\beta$}
\AxiomC{$\Gamma\vdash v:\gamma$}
\RightLabel{(CUT)}
\BinaryInfC{$f_c(f_c(\Phi,\Gamma),\Delta)\vdash u[y:=v]:\beta$}
\DisplayProof
\end{minipage}
\caption{Règles du calcul du système intermédiaire sans élimination $\m{S}_{IC}'$\label{fig:S_IC}}
\end{figure}

\begin{theor}
\label{theor:IE-IC}
Les langages engendrés par $\m{S}_{IE}$ et $\m{S}_{IC}$ sont les mêmes.
\end{theor}

L'annexe \ref{an:IE-IC} contient une preuve de ce premier théorème. En annexe \ref{an:G_0_IC}, on retrouve l'exemple 0 ayant subi cette transformation.

\subsection{Suppression des introductions}

Supprimons maintenant les introductions. L'idée est de se placer à une plus haute introduction (c'est-à-dire sans introduction au-dessus)

\AxiomC{$\vert$}
\noLine
\UnaryInfC{$f_c(\Gamma,x:\alpha)\vdash u:\beta$}
\RightLabel{($c$ I)}
\UnaryInfC{$\Gamma\vdash \uabs{x}{u}:c(\alpha,\beta)$}
\DisplayProof

et de \textbf{réécrire} en
\AxiomC{$\brokenvert$}
\noLine
\UnaryInfC{$\Gamma\vdash \uabs{x}{u}:c(\alpha,\beta)$}
\DisplayProof
la preuve au-dessus (réécriture noté $\rightsquigarrow_1$) \textbf{par induction}. L'hérédité consiste à étudier la forme de la coupure. Dans le premier cas, l'hypothèse de récurrence suffit. Mais il ne faut pas oublier le second cas, où $\Delta$ est vide et $\alpha$ vas dans la prémisse de droite (car $f_c(f_c(\Phi,\Gamma),x:\alpha)=f_c(\Phi,f_c(\Gamma,x:\alpha))$). Cela est résumé en figure [\ref{fig:cas}]. Ce dernier cas nécessite de considérer un autre type de réécriture (noté $\rightsquigarrow_2$) sur la prémisse de droite.

\begin{figure}[ht]
\hspace{-1cm}
\begin{minipage}{9cm}
\centering
\AxiomC{|}
\noLine
\UnaryInfC{$f_c(f_c(f_c(\Phi,y:\gamma),\Delta),x:\alpha)\vdash u:\beta$}
\AxiomC{|}
\noLine
\UnaryInfC{$\Gamma\vdash v:\gamma$}
\RightLabel{(CUT)}
\BinaryInfC{$f_c(f_c(f_c(\Phi,\Gamma),\Delta),x:\alpha)\vdash u[y:=v]:\beta$}
\DisplayProof

\textbf{Cas 1}
\end{minipage}
\begin{minipage}{9cm}
\centering
\AxiomC{|}
\noLine
\UnaryInfC{$f_c(\Phi,y:\gamma)\vdash u:\beta$}
\AxiomC{|}
\noLine
\UnaryInfC{$f_c(\Gamma,x:\alpha)\vdash v:\gamma$}
\RightLabel{(CUT)}
\BinaryInfC{$f_c(f_c(\Phi,\Gamma),x:\alpha)\vdash u[y:=v]:\beta$}
\DisplayProof

\textbf{Cas 2}
\end{minipage}
\caption{Cas de la forme d'une coupure pour la suppression d'une introduction\label{fig:cas}}
\end{figure}

Formellement, le résultat qu'on cherche à montrer est le suivant :

\begin{lem}
\label{lem:elim_intro}
Si $f_c(\Gamma_0,x:\alpha)\vdash u_0:\beta$ est dérivable sans introduction et $u_0$ ne contient que des variables strictement positives (c'est-à-dire dans $\Tp(\m{G})^{+,\neq}$), alors $\Gamma_0\vdash \uabs{x}{u_0}:c(\alpha,\beta)$ (par $\rightsquigarrow_1$) et $f_c(\Gamma_0,z:c(\delta,\alpha))\vdash \uabs{x'}{u_0[x:=\app{z}{x'}]}:c(\delta,\beta)$ (par $\rightsquigarrow_2$) le sont sans introduction (par une dérivation de longueur inférieure, avec des variables aussi strictement positives).
\end{lem}

Il constitue le cœur du problème et sa preuve est disponible en annexe \ref{an:elim_intro}. Notamment, pour le cas de base, on a besoin de générer des axiomes supplémentaires à l'aide des \textbf{opérateurs} suivants, définis pour $\m{A}$ un ensemble de séquents:

\[ \m{Q}_1(\m{A})\triangleq\{\Gamma\vdash  \uabs{x}{u[y:=v]}: c(\alpha,\beta)\ |\ f_c(\Gamma,y:\gamma)\vdash u:\beta\in\m{A}, x:\alpha\vdash_{IE}v:\gamma,\alpha\in\Tp(\m{G})^{+,\neq}\}\]
\[ \m{Q}_2(\m{A})\triangleq\{f_c(\Gamma,z:c(\delta,\alpha))\vdash  \uabs{x}{u[y:= \app{z}{x}]}: c(\delta,\beta)\ |\ f_c(\Gamma,y:\alpha)\vdash u:\beta\in\m{A}\}\]
correspondant à $\rightsquigarrow_1$ et $\rightsquigarrow_2$ respectivement. On peut alors définir le système final $\m{S}_C$ (en figure [\ref{fig:S_C}]) à partir de l'ensemble d'axiomes propres (potentiellement infini) défini par induction:
\[ \m{A}_2::=\m{A}_0~|~\m{Q}_1(\m{A}_2)~|~\m{Q}_2(\m{A}_2) \]

\begin{figure}[ht]
\centering
\begin{minipage}{5cm}
\AxiomC{$\Gamma\vdash u:\alpha~\in\m{A}_2$}
\RightLabel{(pAx.$_{I}$)}
\UnaryInfC{$\Gamma\vdash u:\alpha$}
\DisplayProof
\end{minipage}
\begin{minipage}{6cm}
\AxiomC{$f_c(f_c(\Phi,y:\gamma),\Delta)\vdash u:\beta$}
\AxiomC{$\Gamma\vdash v:\gamma$}
\RightLabel{(CUT)}
\BinaryInfC{$f_c(f_c(\Phi,\Gamma),\Delta)\vdash u[y:=v]:\beta$}
\DisplayProof
\end{minipage}
\caption{Règles du calcul du système final avec seulement la coupure $\m{S}_{C}$\label{fig:S_C}}
\end{figure}

On a  alors le résultat suivant prouvé en annexe \ref{an:IE-C} :

\begin{theor}
\label{theor:IE-C}
Les langages engendrés par $\m{S}_{IE}$ et $\m{S}_{C}$ sont les mêmes.
\end{theor}

Des exemples de traduction vers $\m{S}_C$ sont présents en annexe \ref{an:G_0_IC} pour $\m{G}_0$ et \ref{an:G_1} pour $\m{G}_1$.

\subsection{Des coupures aux ACG}

Ayant tout transformé en coupure, on peut maintenant mettre en place notre (presque\footnote{'presque' car le nombre de symboles non terminaux (et de productions) peut être infini}) grammaire algébrique définie en figure [\ref{CFG}]. Le théorème suivant découle facilement (en annexe \ref{an:C-CFG}) :

\begin{theor}
\label{theor:C-CFG}
$\m{L}(\m{G})=\m{L}(\m{G}_A)$
\end{theor}

\begin{figure}[ht]
$\m{G}_A\triangleq (N,\m{T},P,s)$ où 
\[ N\triangleq\bigcup_{\footnotesize\begin{array}{c}
f_{c_n}(...f_{c_1}(h,\alpha_1),...\alpha_n)\vdash\beta\in\m{A}_2\\
h\in\m{T}\text{ ou }h\in\Tp(\m{G})^{+,\neq}
\end{array}}\{\alpha_1,...\alpha_n,h,\beta\} \]
et 
\[ P\triangleq \{\beta\to\Gamma~|~\Gamma\vdash\beta\in\m{A}_2\}\]
\caption{Grammaire algébrique $\m{G}_A$ à partir de la grammaire de Lambek $\m{G}$ \label{CFG}}
\end{figure}

Enfin, il nous reste à traduire une grammaire algébrique $\m{G}_A=(N,A_0,P,s)$ en ACG. On commence par poser $\Sigma_1^A\triangleq(N,P,(K,w)\mapsto \llbracket w\rrbracket_K)$ avec la définition par induction :
\[ \begin{array}{cl}
\llbracket \varepsilon\rrbracket_K=K & \\
\llbracket Lw\rrbracket_K =L\to\llbracket w\rrbracket_K & \text{si $L\in N$} \\
\llbracket aw\rrbracket_K = \llbracket w\rrbracket_K & \text{si $a\in A_0$}
\end{array} \]
Le vocabulaire objet est celui des chaînes de caractères $\Sigma_2^A\triangleq(\{*\},A_0,a\mapsto\sigma)$, et le lexique $\mathfrak{L}^A\triangleq(N\mapsto\sigma,G^A)$, où $G^A(K,w)=\uabs{\mathbf{x}}{|w|}$ avec $\mathbf{x}$ la suites des variables libres de $|w|$ défini par
\[\begin{array}{cl}
|\varepsilon|=\uabs{x}{x} & \\
|Lw|= y+|w| & \text{si $L\in N$, avec $y$ est une variable fraîche} \\
|aw|= a+|w| & \text{si $a\in A_0$}
\end{array}\]

On a construit $\mathfrak{G}_A\triangleq(\Sigma_1^A,\Sigma_2^A,\mathfrak{L}^A,s)$. S'ensuit le théorème suivant, dont un preuve peut être trouvée chez \cite{dGPog04}.

\begin{theor}
$\mathfrak{O}(\mathfrak{G}^A)=\m{L}(\m{G})$ et $\mathfrak{A}(\mathfrak{G}^A)$ est isomorphe à l'ensemble des arbres de dérivation de $G$.
\end{theor}

On a donc réussi à traduire une grammaire de Lambek en grammaire catégorielle abstraite.

\section{Algorithme de mise à niveau}

\subsection{Mise à niveau}
\label{sec:ex_non_pert}

Le (seul) problème de la traduction précédente est qu'elle peut générer un nombre infini de symboles non terminaux et de productions. Il ne s'agit donc pas vraiment d'une grammaire algébrique (ni donc de d'une ACG) à proprement parler. Cependant, la grande astuce de notre approche a été le côté itératif et l'utilisation d'opérateurs, qui génèrent en nombre fini si l'ensemble de départ est fini. En fait, on considère un \textbf{algorithme} [\ref{algo:man}] dit \textbf{de mise à niveau} :

\begin{algorithm}[H]
\DontPrintSemicolon
\SetKwData{B}{B}
\SetKwInOut{Input}{Entrée}\SetKwInOut{Output}{Sortie}
\Input{Un ensemble d'axiomes propres $\m{A}_0$ pour $\m{S}_{IC}$ et un entier $n$}
\KwData{$\B_0,\B_1$ deux ensembles de séquents}
\Output{Un ensemble d'axiomes propres $\B_1$ intermédiaire pour $\m{S}_C$}
\BlankLine

$\B_1\leftarrow\m{A}_0$\;
\emph{Boucle principale}\;
\For{$i\leftarrow 1$ \KwTo $n$}{
	$\B_0\leftarrow\B_1$\;
    $\B_1\leftarrow\B_1\cup\m{Q}_1(\B_0)$\;
    $\B_1\leftarrow\B_1\cup\m{Q}_2(\B_0)$\;
}
\Return{$\B_1$}\;
\caption{Algorithme de mise à niveau (MaN)}\label{algo:man}
\end{algorithm}

En notant $\m{A}_{\FuncSty{MaN}}^i$ la valeur de $\KwSty{B}_1$ après $i$ tours de boucle principale, et $\m{S}_C^i$ le système qui prend ses axiomes dans $\m{A}_{\FuncSty{MaN}}^i$, on peut démontrer le résultat suivant :
\begin{theor}
\label{theor:reecr-imbr}
Les séquents de $\m{S}_C^i$ dérivables sont exactement ceux de $\m{S}_{IE}$ contenant au plus $i$ introductions imbriquées.
\end{theor}
 où la notion d'introductions imbriquées est explicitée dans la sous-section suivante. Le preuve est trop longue pour être détaillée, même en annexe.

De plus, la version présentée ici de l'algorithme [\ref{algo:man}] est simplifiée. En réalité, il est plus intéressant de n'effectuer les opérations que sur les séquents qui peuvent devenir accessibles (au sens de la grammaire algébrique) par ce procédé. On évite ainsi de produire, dans l'exemple 0, $np,\textsc{M},np/n\vdash s/n$ (car aucun axiome propre n'a pour argument $s/n$) qui pourrait donner des dérivations absurdes comme \textsc{(Pierre voit le) chat}, qui sont présentes chez Pentus. Cela raffine beaucoup l'ensemble final produit, donc s'avère bien mieux que la construction de Pentus.

\subsection{Impact des introductions imbriquées et avantages linguistiques}

Pour comprendre en quoi notre approche est avantageuse, analysons d'abord le problème de la génération infinie d'axiomes propres. Fixons-nous une grammaire de Lambek dans laquelle il existe un séquent dérivable et un axiome $\Gamma\vdash\beta$ utilisé dans cette preuve qui, par la réécriture du lemme [\ref{lem:elim_intro}], se voit opéré $i$ fois. Cet axiome est donc "sur le chemin" des $i$ introductions lors de la réécriture, ou on peut encore dire intuitivement qu'il se trouve "entre l'introduction de $x$ et la consommation\footnote{une seule consommation car on est dans des preuves/$\lambda$-termes linéaires} de $x$". Ces introductions sont alors dite imbriquées, définies plus précisément en figure [\ref{fig:imbr}].

\begin{figure}[ht]
On dit que les variables $x_1,...x_n$ (toutes différentes) sont imbriquées, et on note $x_1\succ x_2\succ...\succ x_n$ pour $x_n$ imbriquée dans $x_{n-1}$, dans ... dans $x_1$, si toutes les introductions de ces variables se font sur une même branche, dans cet ordre avec $x_n$ au plus haut, et s'il existe un sous-terme contenant toutes ces variables libres. Autrement dit si la dérivation est de la forme :\vspace{0.5cm}\centering

\AxiomC{|}
\noLine 
\UnaryInfC{$f_{c_n}(\Gamma_n,x_n:\alpha_n)\vdash u_n:\beta_n$}
\RightLabel{($c$ I)}
\UnaryInfC{$\Gamma_n\vdash\uabs{x_n}{u_n}:c_n(\alpha_n,\beta_n)$}
\noLine 
\UnaryInfC{$\vdots$}
\noLine 
\UnaryInfC{$f_{c_1}(\Gamma_1,x_1:\alpha_1)\vdash u_1:\beta_1$}
\RightLabel{($c$ I)}
\UnaryInfC{$\Gamma_1\vdash\uabs{x_1}{u_1}:c_1(\alpha_1,\beta_1)$}
\noLine
\UnaryInfC{|}
\DisplayProof\vspace{0.5cm}

avec $\{x_1,...x_n\}$ étant des variables libres de $u_n$ (ce qui est équivalent à ce cité juste au-dessus).
\caption{Définition de l'imbrication d'introductions\label{fig:imbr}}
\end{figure}

La fait qu'une grammaire de Lambek nécessite une infinité d'applications des opérateurs (exemple en annexe \ref{an:G_1}) est donc équivalent à l'existence de séquents avec un $\lambda$-terme contenant un sous-terme ($u_n$ dans la définition) avec un nombre arbitrairement grand de variables libres. J'ai essayé d'analyser plus finement la condition dans le but de pouvoir la détecter (par exemple par la détection de boucle en fonction de la forme des axiomes propres), mais le temps m'a manqué.

Malgré tout, d'un point de vue linguistique, le besoin d'effectuer une opération $\m{Q}_j$ correspond à l'extraction d'un complément dans un syntagme. On observe le plus souvent ce phénomène dans les relatives, comme le montre l'exemple 0 en annexe \ref{an:G_0_IC}. Or dans le langage naturel, les relatives sont des "îlots à l'extraction", c'est-à-dire qu'on ne peut rien extraire d'une relative. Par exemple, on ne peut rien extraire de la proposition relative dans la phrase "Pierre qui voit le chat". Notamment, on ne peut pas dire "le chat que Pierre qui voit". C'est pourquoi le phénomène d'imbrication d'introductions  arbitraire est hautement improbable dans le cadre où l'on se restreint, d'où la correction totale de l'algorithme [\ref{algo:man}] après seulement quelques tours de grande boucle (3 suffisent généralement).

\section{Conclusion}

En somme, après l'analyse d'une construction déjà effective mais insatisfaisante en pratique d'une traduction d'une grammaire de Lambek en grammaire catégorielle abstraite, j'ai développé et démontré la méthode qui consiste à transformer au fur et à mesure les éliminations et introductions en coupure, pour passer par une grammaire algébrique. Ce principe est intéressant car, linguistiquement, il suffit de faire seulement quelques opérations pour obtenir une traduction plus concise et moins ambiguë.

Malgré tout la méthode n'est intéressante que pour les cas linguistiques, et ne saisit pas toute la puissance des grammaires des Lambek. Bien que la première méthode fonctionne toujours, je pense qu'il y a moyen d'approfondir la recherche sur les imbrications d'introduction pour, si possible, engendrer de manière finie.

Ce résultat reste néanmoins pertinent car il montre comment encoder dans les ACG, de manière plus simple que ce qui existait auparavant, une façon de prendre fortement en considération la syntaxe de la phrase. De plus, on peut envisager de l'étendre à des dérivées des grammaires de Lambek plus utilisées de nos jours (car plus complexes). Enfin, cela apporte une méthode supplémentaire au maniement des systèmes de dérivation.


\clearpage

\appendix

\section{Problèmes de la traduction naïve}
\label{an:naif}

Outre le fait que la traduction naïve ne prend en compte l'orientation, le système de typage dans les ACG étant différent, il permet une inversion de l'ordre des arguments qui n'est pas possible chez Lambek, car le membre de gauche est un mot, et non pas un ensemble. Par exemple, pour une grammaire $\m{G}_3$ définie par $A:(p/q)/r$ et $B:s/((p/r)/q)$, les types étant pris atomiques tous différents, on voit bien en essayant de construire la dérivation qu'on ne peut pas prouver $BA\vdash s$. Effectivement, $y$ ne peut pas être sorti car il est entre $A$ et $z$.\vspace{0.5cm}

\AxiomC{$B\vdash B:s/((p/r)/q)$}
\AxiomC{$
z:r,A\vdash\app{A}{z}:p/q$}
\AxiomC{$\cB{y\!:q}\vdash y:q$}
\RightLabel{Impossible}
\BinaryInfC{$z:r,\cB{y\!:q},A\vdash \app{\app{A}{z}}{y}:p$}
\RightLabel{(/I)}
\UnaryInfC{$y:q,A\vdash \uabs{z}{\app{\app{A}{z}}{y}}:p/r$}
\RightLabel{(/I)}
\UnaryInfC{$A\vdash \uabs{y,z}{\app{\app{A}{z}}{y}}:(p/r)/q$}
\RightLabel{(/E)}
\BinaryInfC{$BA\vdash \app{B}{\uabs{y,z}{\app{\app{A}{z}}{y}}}:s$}
\DisplayProof\vspace{0.5cm}

Alors que le $\lambda$-terme d'ACG $\app{B}{\uabs{y,z}{\app{\app{A}{z}}{y}}}$ est constructible.

\section{Preuve du théorème \ref{theor:IE-IC}}
\label{an:IE-IC}

\begin{proof}\ 
\begin{itemize}
\item[\fbox{$\subseteq$}] Il suffit de montrer que $\m{S}_{IC}$ est admissible dans $\m{S}_{IE}$.

Pour la coupure, il s'agit du résultat de Grenzen [Lam58].

Pour les axiomes propres (pAx.$_{Ej}$), montrons pour $\alpha=c_1(\alpha_1,...c_n(\alpha_n,r)...)$ par récurrence sur $i$ que $f_{c_{i-1}}(...f_{c_1}(h,z_1:\alpha_1),...z_{i-1}:\alpha_{i-1}) \vdash_{IE} \app{h}{z_1...z_{i-1}}:c_i(\alpha_i,...c_n(\alpha_n,r)...)$, avec $h\in\m{T}$ ou\footnote{On note abusivement toujours $h$ pour le type ou le couple variable\!:type} $h\in\Tp(\m{G})^{+,\neq}$.

(I) Pour $i=1$, c'est un axiome (Lex.) ou (Ax.) de $\m{S}_{IE}$ en fonction de $h$.

(H) L'hérédité est directe en appliquant ($c_i$ E) :\vspace{0.5cm}

\AxiomC{$f_{c_{i-1}}(...f_{c_1}(h,z_1:\alpha_1),...z_{i-1}:\alpha_{i-1})  \vdash \app{h}{z_1...z_{i-1}}:c_i(\alpha_i,c_{i+1}(\alpha_{i+1},...c_n(\alpha_n,r)...))$}
\AxiomC{}
\RightLabel{(Ax.)}
\UnaryInfC{$z_i:\alpha_i\vdash z_i:\alpha_i$}
\RightLabel{($c_i$ E)}
\BinaryInfC{$f_{c_i}(f_{c_{i-1}}(...f_{c_1}(h,z_1:\alpha_1),...z_{i-1}:\alpha_{i-1}),z_i:\alpha_i) \vdash \app{\app{h}{z_1...z_{i-1}}}{z_i}:c_{i+1}(\alpha_{i+1},...c_n(\alpha_n,r)...)$}
\DisplayProof\vspace{0.5cm}

\item[\fbox{$\supseteq$}] On cherche à réécrire dans $\m{S}_{IC}$ la preuve $\beta$-normale $\eta$-longue du séquent $\Gamma_0\vdash_{IE} u_0:s$.

\quad Rappelons que tout $\lambda$-terme $u$ peut se mettre sous une \textbf{forme} unique dite \textbf{de tête} $u=\uabs{x_1,...x_m}{\app{h}{u_1...u_n}}$, où $h$ est la tête, et $u$ est dit sous forme normale de tête si $h$ est une constante ou une variable.\vspace{0.5cm}

\quad Avec cet outil, montrons par récurrence sur la dérivation de $\Gamma\vdash u:\beta$ dans $\m{S}_{IE}$, que si la tête $h$ de $u$ est un symbole lexical $h\in\m{T}$ ou une variable telle que $h\in\Tp(\m{G})^{+,\neq}$, et si toutes les variables de $\Gamma$ et les variables $x_i$ introduites par $u$ sont strictement positives, alors toutes les variables de $u$ ont un type strictement positif (c'est-à-dire dans $\Tp(\m{G})^{+,\neq}$).

(I) Si $\Gamma\vdash u:\beta$ est un axiome, $h=\Gamma=u$. Le résultat est alors direct.

(H) Pour $u=\uabs{x_1,...x_m}{\app{h}{u_1...u_n}}$ sous forme de tête, et $h$ de type $\alpha=c_1(\alpha_1,...c_n(\alpha_n,r)...)$. Soit $i\in\llbracket 1,n\rrbracket$. Notons $u_i=\uabs{x_1^i,...x_{m_i}^i}{\app{h^i}{u_1^i...u_{n_i}^i}}$. Prenons $u^i$ du type $\alpha_i=c_1^i(\alpha_1^i,...c_{n_i}^i(\alpha_{n_i}^i,r^i)...)$. Puisque $\alpha$ est positif par hypothèse, $\alpha_i\subset^-\alpha$ et $\alpha_j^i \subset^-\alpha_i$ donnent $\alpha_j^i\in\Tp(\m{G})^{+,\neq}$, donc les variables introduites par $u_i$ sont strictement positives. Et de même pour celles de $\Gamma_i$ tel que $\Gamma_i\vdash u_i:\alpha_i$, puisqu'elles sont parmi celles de $\Gamma$ et les $x_j$, strictement positives par hypothèse. Pour la tête de $u_i$, on a cependant 3 cas : $h^i\in\m{T}$, $h^i$ est une variable de $\Gamma_i$ et $h^i$ est une variable $x_j^i$. Dans tous les cas, par hypothèse $h^i$ est de type strictement positif, on peut appliquer l'hypothèse d'induction et alors toutes les variables de $u_i$ sont strictement positives. Puisque par hypothèse c'est déjà le cas pour celles introduites par $u$, c'est donc le cas pour $u$ aussi.

\quad Revenons au séquent premier $\Gamma_0\vdash u_0:s$. Puisque $s$ est atomique, $u_0$ n'introduit pas de variables ($m=0$, donc elles sont toutes strictement positives). Puisque $\Gamma\in\m{T}^+$ (donc toutes ses variables sont dans $\Tp(\m{G})^{+,\neq}$), la tête $h$ ne peut être qu'une constante. On peut donc utiliser le résultat précédent, et toutes les variables de $u_0$ sont bien strictement positives.\vspace{0.5cm}

\quad On peut passer au résultat principal. Montrons par induction sur la dérivation de $\Gamma\vdash u:\beta$ dans $\m{S}_{IE}$ qu'on peut réécrire ce séquent dans $\m{S}_{IC}$, si toutes les variables de $u$ sont strictement positives. L'inclusion souhaitée s'ensuivra directement.

(I) Si la dernière règle est (Ax.) sur $x:\alpha\vdash x:\alpha$, alors par hypothèse $x$ est une occurrence positive, donc on peut réécrire avec la règle (pAx.$_{E2}$). Si la dernière règle est (Lex.) sur $t\vdash t:\alpha=r$, avec $r$ atomique, la règle (pAx.$_{E1}$) convient. Si $\alpha$ n'est pas atomique, puisqu'on prend la preuve $\eta$-longue, le lexème est appliqué à des arguments, c'est donc le cas d'hérédité qui le traite.

(H) Prenons $u=\uabs{x_1,...x_m}{\app{h}{u_1...u_n}}$ sous sa forme de tête,\ $\beta=c_1(\beta_1,...c_n(\beta_m,r)...)$ ($r$ atomique) et $h$ de type $\alpha=c_1'(\alpha_1,...c_n'(\alpha_n,r)...)$. D'après le $\lambda$-terme, les dernières règles de la preuve sont $m$ introductions jusqu'à $\Xi=f_{c_m}(...f_{c_1}(\Gamma,x_1:\beta_1,),...x_m:\beta_m)\vdash \app{h}{u_1...u_n}:r$. Et encore au-dessus, $n$ éliminations de $\Xi=f_{c_n'}(...f_{c_1'}(h,\Delta_1),...\Delta_n)\vdash \app{h}{u_1...u_n}:r$ vers $h\vdash h:c_1'(\alpha_1,...c_n'(\alpha_n,r)...)$, laissant à prouver les séquents $\Delta_i\vdash u_i:\alpha_i$ de la même forme que $\Gamma\vdash u:\beta$ et ne contenant que des variables positives. La dérivation décrite est dépeinte dans la figure \ref{derIE}.

\begin{figure}[ht]
\AxiomC{}
\RightLabel{(Ax.) ou (Lex.)}
\UnaryInfC{$h\vdash h:c_1'(\alpha_1,...c_n'(\alpha_n,r)...)$}
\AxiomC{|}
\noLine
\UnaryInfC{$\Delta_1\vdash u_1:\alpha_1$}
\RightLabel{($c_1'$ E)}
\BinaryInfC{$\vdots$}
\noLine
\UnaryInfC{$f_{c_{n-1}'}(...f_{c_1'}(h,\Delta_1),...\Delta_{n-1})\vdash \app{h}{u_1...u_{n-1}}:c_n'(\alpha_n,r)$}
\AxiomC{|}
\noLine
\UnaryInfC{$\Delta_n\vdash u_n:\alpha_n$}
\RightLabel{($c_n'$ E)}
\BinaryInfC{$f_{c_n'}(...f_{c_1'}(h,\Delta_1),...\Delta_n)=\Xi\vdash \app{h}{u_1...u_n}:r$}
\noLine
\UnaryInfC{$\shortparallel$}
\noLine
\UnaryInfC{$f_{c_m}(...f_{c_1}(\Gamma,x_1:\beta_1),...x_m:\beta_m)\vdash \app{h}{u_1...u_n}:r$}
\RightLabel{($c_m$ I)}
\UnaryInfC{$\vdots$}
\noLine
\UnaryInfC{$f_{c_1}(\Gamma,x_1:\alpha_1)\vdash \uabs{x_2,...x_m}{\app{h}{u_1...u_n}}:c_2(\beta_2,...c_m(\beta_m,r)...)$}
\RightLabel{($c_1$ I)}
\UnaryInfC{$\Gamma\vdash u:\beta=c_1(\beta_1,...c_m(\beta_m,r)...)$}
\DisplayProof
\caption{Preuve $\beta$-normale $\eta$-longue de $\Gamma\vdash u:\beta$ dans $\m{S}_{IE}$\label{derIE}}
\end{figure}

On réécrit cette  dérivation en remplaçant les éliminations par des coupures. Pour cela, on utilise l'hypothèse de récurrence sur les $\Delta_i\vdash u_i:\beta_i$. À partir de $\Xi\vdash \app{h}{u_1...u_n}:r$, on garde la fin de la dérivation. Le début de la preuve réécrite est visible dans la figure \ref{derIC}.

\begin{figure}[ht]
\AxiomC{}
\RightLabel{(pAx.$_{Ej}$)}
\UnaryInfC{$f_{c_n'}(...f_{c_1'}(h,z_1:\alpha_1),...z_n:\alpha_n)\vdash \app{h}{z_1...z_n}:r$}
\AxiomC{$\brokenvert$}
\noLine
\UnaryInfC{$\Delta_1\vdash u_1:\alpha_1$}
\RightLabel{(CUT)}
\BinaryInfC{$\vdots$}
\noLine
\UnaryInfC{$f_{c_{n-1}'}(...f_{c_1'}(h,\Delta_1),...z_n:\beta_n)\vdash \app{\app{h}{u_1...u_{n-1}}}{z_n}:r$}
\AxiomC{$\brokenvert$}
\noLine
\UnaryInfC{$\Delta_n\vdash u_n:\alpha_n$}
\RightLabel{(CUT)}
\BinaryInfC{$f_{c_n'}(...f_{c_1'}(h,\Delta_1),...\Delta_n)=\Xi\vdash \app{h}{u_1...u_n}:r$}
\DisplayProof
\caption{Haut de la preuve réécrite de $\Gamma\vdash u:\beta$ dans $\m{S}_{IC}$\label{derIC}}
\end{figure}

Ainsi tout mot $\Gamma_0$ tel que $\Gamma_0\vdash_{IE}u_0:s$ est aussi prouvable dans $\m{S}_{IC}$.

\end{itemize}
\end{proof}

Remarquons qu'on a même plus fort : les dérivations sont conservées, c'est-dire que si $\Gamma_0\vdash_{IE}u_0:s$ alors $\Gamma_0\vdash_{IC}u_0:s$ (le $\lambda$-terme est identique). L'équivalence est donc forte.

\section{Traduction de l'exemple 0 sans élimination, puis sans introduction}
\label{an:G_0_IC}

Voici la dérivation de l'exemple 0 traduite dans $\m{S}_{IC}$. Pour tout placer sur une largeur, on se contente du sous-séquent $\textsc{chat,que,Pierre,voit} \vdash n$, le reste est analogue.\vspace{0.5cm}

\hspace{-1cm}
{\footnotesize
\AxiomC{$z_2:n,\textsc{q},z_1:s/np\vdash \app{\app{\textsc{q}}{z_1}}{z_2}:n$}
	\AxiomC{$z_2:np,\textsc{v},z_1:np\vdash \app{\app{\textsc{v}}{z_1}}{z_2}:s$}
	\AxiomC{$x:np\vdash x:np$}
	\RightLabel{(CUT)}
	\BinaryInfC{$z_2:np,\textsc{v},x:np\vdash \app{\app{\textsc{v}}{x}}{z_2}:s$}
	\AxiomC{$\textsc{P}\vdash\textsc{P}:np$}
	\RightLabel{(CUT)}
	\BinaryInfC{$\textsc{P,v},x:np\vdash \app{\app{\textsc{v}}{x}}{\textsc{P}}:s$}
\RightLabel{(/I)}
\UnaryInfC{$\textsc{P,v}\vdash \uabs{x}{\app{\app{\textsc{v}}{x}}{\textsc{P}}}:s/np$}
\RightLabel{(CUT)}
\BinaryInfC{$z_2:n,\textsc{q,P,v}\vdash \app{\app{\textsc{q}}{\uabs{x}{\app{\app{\textsc{v}}{x}}{\textsc{P}}}}}{z_2}:n$}
\AxiomC{$\textsc{c}\vdash\textsc{c}:n$}
\RightLabel{(CUT)}
\BinaryInfC{$\textsc{c,q,P,v}\vdash \app{\app{\textsc{q}}{\uabs{x}{\app{\app{\textsc{v}}{x}}{\textsc{P}}}}}{\textsc{c}}:n$}
\DisplayProof
}\vspace{0.5cm}

Et voici le résultat en supprimant l'introduction. Cela demande l'utilisation de l'axiome propre opéré (par $\m{Q}_1$) $z_2:np,\textsc{v}\vdash\uabs{x}{\app{\app{\textsc{v}}{x}}{z_1}}:s/np$ où $z_1$ a été $\alpha$-renommée en $x$ (très proche de la version via Pentus \textsc{voit$_0$}$:\ch{np}\to\ch{s/np}$).\vspace{0.5cm}

\hspace{-1cm}
{\footnotesize
\AxiomC{$z_2:n,\textsc{q},z_1:s/np\vdash \app{\app{\textsc{q}}{z_1}}{z_2}:n$}
	\AxiomC{$z_2:np,\textsc{v}\vdash \uabs{x}{\app{\app{\textsc{v}}{x}}{z_2}}:s$}
	\AxiomC{$\textsc{P}\vdash\textsc{P}:np$}
	\RightLabel{(CUT)}
\BinaryInfC{$\textsc{P,v}\vdash \uabs{x}{\app{\app{\textsc{v}}{x}}{\textsc{P}}}:s/np$}
\RightLabel{(CUT)}
\BinaryInfC{$z_2:n,\textsc{q,P,v}\vdash \app{\app{\textsc{q}}{\uabs{x}{\app{\app{\textsc{v}}{x}}{\textsc{P}}}}}{z_2}:n$}
\AxiomC{$\textsc{c}\vdash\textsc{c}:n$}
\RightLabel{(CUT)}
\BinaryInfC{$\textsc{c,q,P,v}\vdash \app{\app{\textsc{q}}{\uabs{x}{\app{\app{\textsc{v}}{x}}{\textsc{P}}}}}{\textsc{c}}:n$}
\DisplayProof
}\vspace{0.5cm}

\section{Preuve du lemme \ref{lem:elim_intro}}
\label{an:elim_intro}

\begin{proof}
On considère la preuve $\beta$-normale $\eta$-longue de $u_0$, et on va la réécrire  sans introduction (noté $\rightsquigarrow_1$ et $\rightsquigarrow_2$ respectivement), par récurrence sur la hauteur de la dérivation.

(I) Si la dernière règle est un axiome, alors par clôture par $\m{Q}_1$ (donnant $\Gamma_0\vdash \uabs{x}{u_0[x:=x]}:c(\alpha,\beta)$ en prenant $v=x$ et $\gamma=\alpha$) et $\m{Q}_2$ (donnant $f_c(\Gamma_0,z:c(\delta,\alpha))\vdash \uabs{x'}{u_0[x:=\app{z}{x'}]}:c(\delta,\beta)$), ces séquents sont dans $\m{A}_2$ donc on peut utiliser (pAx.$_I$). Remarquons que pour $\rightsquigarrow_1$, on traite la la tête qui est la variable $x:\alpha$. 

(H) Si la dernière règle est une coupure, il y a deux cas, dont un seulement si $\Delta=\emptyset$.

\textbf{Cas 1 : } $x:\alpha$ va dans la prémisse de gauche. On peut alors utiliser l'hypothèse de récurrence sur cette prémisse (figure \ref{fig:elim_intro_1}). Remarquons qu'il n'y a pas de problème pour inverser les substitutions dans la cas \textbf{1.2} car $z$ et $x'$ sont prises fraîches, et qu'on travaille avec des $\lambda$-termes linéaires, donc $x$ n'apparaît pas dans $v$.

\begin{figure}[ht] \footnotesize
\hspace{-1cm}{
$\begin{array}{rccr}
\AxiomC{|}
\noLine
\UnaryInfC{$f_c(f_c(f_c(\Phi,y:\gamma),\Delta),x:\alpha)\vdash u:\beta$}
\AxiomC{|}
\noLine
\UnaryInfC{$\Gamma\vdash v:\gamma$}
\RightLabel{(CUT)}
\BinaryInfC{$f_c(f_c(f_c(\Phi,\Gamma),\Delta),x:\alpha)\vdash u[y:=v]:\beta$}
\DisplayProof
& \rightsquigarrow_1 &
\AxiomC{$\brokenvert_1$}
\noLine
\UnaryInfC{$f_c(f_c(\Phi,y:\gamma),\Delta)\vdash\uabs{x}{u}:c(\alpha,\beta)$}
\AxiomC{|}
\noLine
\UnaryInfC{$\Gamma\vdash v:\gamma$}
\RightLabel{(CUT)}
\BinaryInfC{$f_c(f_c(\Phi,\Gamma),\Delta)\vdash\uabs{x}{u[y:=v]}:c(\alpha,\beta)$}
\DisplayProof
& \textbf{Cas 1.1}
\end{array}$}
\vspace{0.3cm}

\centering
$\begin{array}{cr}
\rightsquigarrow_2
\AxiomC{$\brokenvert_2$}
\noLine
\UnaryInfC{$f_c(f_c(f_c(\Phi,y:\gamma),\Delta),z:c(\delta,\alpha))\vdash\uabs{x'}{u[x:=\app{z}{x'}]}:c(\delta,\beta)$}
\AxiomC{|}
\noLine
\UnaryInfC{$\Gamma\vdash v:\gamma$}
\RightLabel{(CUT)}
\BinaryInfC{$f_c(f_c(f_c(\Phi,\Gamma),\Delta),z:c(\delta,\alpha))\vdash \uabs{x'}{u[x:=\app{z}{x'},y:=v]}:c(\delta,\beta)$}
\DisplayProof
& \textbf{Cas 1.2}
\end{array}$
\caption{Réécriture de la preuve pour le lemme \ref{lem:elim_intro} dans le \textbf{cas 1}\label{fig:elim_intro_1}}
\end{figure}

\textbf{Cas 2 : } $x:\alpha$ va dans l'hypothèse de droite (seulement si $\Delta=\emptyset$). On utilise alors la bonne hypothèse de récurrence sur les deux prémisses.

\begin{figure}[ht] \footnotesize
\hspace{-1.7cm}{
$\begin{array}{rccl}
\AxiomC{|}
\noLine
\UnaryInfC{$f_c(\Phi,y:\gamma)\vdash u:\beta$}
\AxiomC{|}
\noLine
\UnaryInfC{$f_c(\Gamma,x:\alpha)\vdash v:\gamma$}
\RightLabel{(CUT)}
\BinaryInfC{$f_c(\Phi,f_c(\Gamma,x:\alpha))\vdash u[y:=v]:\beta$}
\DisplayProof
& \rightsquigarrow_1 &
\AxiomC{$\brokenvert_2$}
\noLine
\UnaryInfC{$f_c(\Phi,z:c(\alpha,\gamma))\vdash\uabs{x}{u[y:=\app{z}{x}]}: c(\alpha,\beta)$}
\AxiomC{$\brokenvert_1$}
\noLine
\UnaryInfC{$\Gamma\vdash\uabs{x}{v}: c(\alpha,\gamma)$}
\RightLabel{(CUT)}
\BinaryInfC{$f_c(\Phi,\Gamma)\vdash\uabs{x}{u[y:=\app{\uabs{x}{v}}{x}]}\to_{\beta}\uabs{x}{u[y:=v]}:c(\alpha,\beta)$}
\DisplayProof
& \textbf{Cas 2.1 } (\Gamma\neq\emptyset)
\end{array}$}
\vspace{0.3cm}

\hspace{-0.5cm}
$\begin{array}{cr}
 \rightsquigarrow_2
\AxiomC{$\brokenvert_2$}
\noLine
\UnaryInfC{$f_c(\Phi,z':c(\delta,\gamma))\vdash\uabs{x'}{u[y:=\app{z'}{x'}]}:c(\alpha,\beta)$}
\AxiomC{$\brokenvert_2$}
\noLine
\UnaryInfC{$f_c(\Gamma,z:c(\delta,\alpha))\vdash\uabs{x''}{v[x:=\app{z}{x''}]}:c(\delta,\gamma)$}
\RightLabel{(CUT)}
\BinaryInfC{$f_c(\Phi,f_c(\Gamma,z:c(\delta,\alpha)))\vdash\uabs{x'}{u[y:=\app{\uabs{x''}{v[x:=\app{z}{x''}]}}{x'}]}\to_{\beta}\uabs{x'}{u[y:=v[x:=\app{z}{x'}]]}=\uabs{x'}{(u[y:=v])[x:=\app{z}{x'}]}: c(\alpha,\beta)$}
\DisplayProof
& \textbf{Cas 2.2}
\end{array}$
\caption{Réécriture de la preuve pour le lemme \ref{lem:elim_intro} dans le \textbf{cas 2}\label{fig:elim_intro_2}}
\end{figure}

Dans le cas dégénéré où $\Gamma=\emptyset$, il faut remonter à la tête, appliquer $\m{Q}_1$ et supprimer la coupure. On ne va pas essayer de savoir exactement quelle forme a la dérivation jusque là, mais on va juste prendre ce qui ressemble le plus à la traduction dans $\m{S}_{IC}$. Dans tous les cas, on peut extraire le fil le la dérivation en remontant par la prémisse gauche à chaque fois, jusqu'à l'axiome propre $\Gamma_1\vdash u_1:\beta$, de variables de coupure $y_1:\gamma_1,...y_n:\gamma_n$ dans $\Gamma_1$ et $u_1$, coupant respectivement sur $\Delta_i\vdash v_i:\gamma_i$, formant par récurrence $\Gamma_{i+1}\triangleq\Gamma_{i}[y_i:=v_i]$ et $u_{i+1}=u_i[y_i:=v_i]$ (de telle sorte que $\Gamma_0=\Gamma_{n+1}$ et $u_0=u_{n+1}$).

Or, on a que $x:\alpha$ est sur le bord du mot $\Gamma_0$. En effet, puisque qu'à l'introduction de $x:\alpha$, il est sur un bord du mot à gauche du séquent, et que les coupures n'ajoutent pas de lettres sur les bord, ou modifient directement la lettre du bord (ce qui n'est pas le cas de $x:\alpha$ qui n'est pas une variable de coupure), $\Delta_n=x:\alpha$ est donc restée sur le bord. Par ailleurs, $x\in\Tp(\m{G})^{+,\neq}$ car toutes les variables sont strictement positives par hypothèse.

Il est alors possible de réécrire $\Gamma_1=f_c(\Gamma_1',y_n:\gamma_n)\vdash u_1:\beta$ par $\m{Q}_1$ en $\Gamma_1'\vdash u_1':c(\alpha,\beta)$ (où $u_1'\triangleq \uabs{x}{u_1[y_n:=v_n]}$), et même toute la preuve avec $\Gamma_i=f_c(\Gamma_i',y_n:\gamma_n)$ (donc $\Gamma_n'=\Gamma_0'$), et $u_{i+1}'\triangleq u_i'[y_i:=v_i]$ pour $i<n$. Par la fraîcheur des $y_i$ de coupure et la linéarité des $\lambda$-termes, aucun $y_i$ (ni $x$) n'apparaît dans un $v_j$ (et sont différents de $x$). On peut ainsi échanger les substitutions de telle sorte que $u_n'=\uabs{x}{u_{n+1}}=\uabs{x}{u_0}$. D'où le séquent souhaité. On a représenté cela en figure [\ref{fig:cas_2.1}]

\begin{figure}[ht]
\centering
\AxiomC{}
\RightLabel{(pAx.$_I$)}
\UnaryInfC{$\Gamma_1=f_c(\Gamma_1',y_n:\gamma_n)\vdash u_1:\beta$}
\AxiomC{|}
\noLine
\UnaryInfC{$\Delta_1\vdash v_1:\gamma_1$}
\RightLabel{(CUT)}
\BinaryInfC{$\Gamma_2=\Gamma_1[y_1:=\Delta_1]\vdash u_2=u_1[y_1:=v_1]:\beta$}
\noLine
\UnaryInfC{$\vdots$}
\noLine 
\UnaryInfC{$\Gamma_n=f_c(\Gamma_n',y_n:\gamma_n)\vdash u_n:\beta$}
\AxiomC{|}
\noLine
\UnaryInfC{$\Delta_n=x:\alpha\vdash v_n:\gamma_n$}
\RightLabel{(CUT)}
\BinaryInfC{$\Gamma_0=f_c(\Gamma_0',x:\alpha)\vdash u_0=u_n[y_n:=v_n]:\beta$}
\DisplayProof\vspace{0.5cm}

\rotatebox{-90}{$\rightsquigarrow$}\vspace{0.5cm}

\AxiomC{}
\RightLabel{(pAx.$_I$)}
\UnaryInfC{$\Gamma_1'\vdash u_1'=\uabs{x}{u_1[y_n:=v_n]}:c(\alpha,\beta)$}
\AxiomC{|}
\noLine
\UnaryInfC{$\Delta_1\vdash v_1:\gamma_1$}
\RightLabel{(CUT)}
\BinaryInfC{$\Gamma_2'\vdash u_2'=u_1'[y_1:=v_1]:c(\alpha,\beta)$}
\noLine
\UnaryInfC{$\vdots$}
\noLine 
\UnaryInfC{$\Gamma_{n-1}'\vdash u_{n-1}':c(\alpha,\beta)$}
\AxiomC{|}
\noLine
\UnaryInfC{$\Delta_{n-1}\vdash v_{n-1}:\gamma_{n-1}$}
\RightLabel{(CUT)}
\BinaryInfC{$\Gamma_0'\vdash \uabs{x}{u_1[y_n:=v_n,y_1:=v_1,...y_{n-1}:=v_{n-1}]}=\uabs{x}{u_0}:c(\alpha,\beta)$}
\DisplayProof
\caption{Réécriture de la preuve dans le cas dégénéré 2.1 du lemme \ref{lem:elim_intro}\label{fig:cas_2.1}}
\end{figure}

Dans tous les cas, on a donc transformé la preuve en un $\lambda$-terme qui se $\beta$-réduit en au plus une étape en celui originel. Ainsi, à $\beta$-équivalence près, la réécriture conserve la structure.

\end{proof}

\section{Preuve du théorème \ref{theor:IE-C}}
\label{an:IE-C}

\begin{proof}\ 

\begin{itemize}
\item[\fbox{$\subseteq$}] La coupure étant admissible dans $\m{S}_{IE}$, il suffit de montrer que les axiomes (pAx.$_{I}$) sont admissibles dans $\m{S}_{IC}$. Par récurrence :

(I) Par un axiome (pAx$_{Ej}$), admissible dans $\m{S}_{IE}$ d'après le théorème \ref{theor:IE-IC}. 

(H) Pour $\m{Q}_1$, on se ramène à l'hypothèse de récurrence via l'arbre :\vspace{0.5cm}

\AxiomC{$\brokenvert$}
\noLine 
\UnaryInfC{$f_c(\Gamma,y:\gamma)\vdash u:\beta$}
\AxiomC{|}
\noLine 
\UnaryInfC{$x:\alpha\vdash v:\gamma$}
\RightLabel{(CUT)}
\BinaryInfC{$f_c(\Gamma,x:\alpha)\vdash u[y:=v]:\beta$}
\RightLabel{($c$ I)}
\UnaryInfC{$\Gamma\vdash \uabs{x}{u[y:=v]}:c(\alpha,\beta)$}
\DisplayProof\vspace{0.5cm}

puisque $x:\alpha\vdash_{IE}v:\gamma$. En pratique, on a souvent $\gamma=\alpha$ (et alors $v=x$) ou des variantes de $\gamma=c(\tilde{c}(\alpha,\delta),\delta)$ (et alors $v=\uabs{g}{\app{g}{x}}$). Pour $\m{Q}_2$, on utilise l'arbre suivant :\vspace{0.5cm}

\AxiomC{$\brokenvert$}
\noLine
\UnaryInfC{$f_c(\Gamma,y:\alpha)\vdash u:\beta$}
\AxiomC{}
\RightLabel{(Ax.)}
\UnaryInfC{$z:c(\delta,\alpha)\vdash z:c(\delta,\alpha)$}
\AxiomC{}
\RightLabel{(Ax.)}
\UnaryInfC{$x:\delta\vdash x:\delta$}
\RightLabel{($c$ E)}
\BinaryInfC{$f_c(z:c(\delta,\alpha),x:\delta)\vdash \app{z}{x}:\alpha$}
\RightLabel{(CUT)}
\BinaryInfC{$f_c(\Gamma,f_c(z:c(\delta,\alpha),x:\delta))= f_c(f_c(\Gamma,z:c(\delta,\alpha)),x:\delta) \vdash u[y:=\app{z}{x}]:\beta$}
\RightLabel{($c$ I)}
\UnaryInfC{$f_c(\Gamma,z:c(\delta,\alpha))\vdash \uabs{x}{u[y:=\app{z}{x}]}:c(\delta,\beta)$}
\DisplayProof\vspace{0.5cm}

Notons que l'associativité $f_c(\Gamma,f_c(c(\delta,\alpha),\delta))= f_c(f_c(\Gamma,c(\delta,\alpha)),\delta)$ est décisive ici.

\item[\fbox{$\supseteq$}] Pour une dérivation $\beta$-normale $\eta$-longue du séquent, commençons par la traduire dans $\m{S}_{IC}$. On utilise alors le lemme précédent sur toutes les introductions successivement, en commençant par la plus haute.

\begin{figure}[ht]
\[\begin{array}{c}
\AxiomC{|}
\noLine
\UnaryInfC{$f_c(\Gamma,x:\alpha)\vdash u:\beta$}
\RightLabel{($c$ I)}
\UnaryInfC{$\Gamma\vdash \uabs{x}{u}:c(\alpha,\beta)$}
\DisplayProof
\rightsquigarrow
\AxiomC{$\brokenvert_1$}
\noLine
\UnaryInfC{$\Gamma\vdash \uabs{x}{u}:c(\alpha,\beta)$}
\DisplayProof
\end{array}\]
\end{figure}

Le $\lambda$-terme obtenu est en fait $\beta$-équivalent à celui de la preuve dans $\m{S}_{IE}$, donc la transformation conserve la structure.

\end{itemize}
\end{proof}

\section{Preuve du théorème \ref{theor:C-CFG}}
\label{an:C-CFG}

\begin{proof} On utilise bien sûr le système $\m{S}_C$
\begin{itemize}
\item[\fbox{$\subseteq$}] Montrons par induction sur la preuve que si $\Gamma_0\vdash_C\beta$ alors $\beta\Rightarrow^*\Gamma_0$.

(I) Si c'est un axiome, il appartient à $P$ donc est dérivable en 1 étape.

(H) Pour
\AxiomC{$f_c(f_c(\Phi,\gamma),\Delta)\vdash\beta$}
\AxiomC{$\Gamma\vdash\gamma$}
\RightLabel{(CUT)}
\BinaryInfC{$\Gamma_0=f_c(f_c(\Phi,\Gamma),\Delta)\vdash\beta$}
\DisplayProof
la dernière règle appliquée, par hypothèse d'induction sur les deux prémisses, $\beta\Rightarrow^* f_c(f_c(\Phi,\gamma),\Delta)\Rightarrow^* f_c(f_c(\Phi,\Gamma),\Delta)$.

D'où le résultat pour $\beta=s$..

\item[\fbox{$\supseteq$}] Montrons par récurrence sur $n$ que si $\beta\Rightarrow^n\Gamma_0$ alors $\Gamma_0\vdash_C\beta$.

(I) Si $n=1$, alors la dérivation est dans $P$ donc est un axiome propre.

(H) Si la dernière règle appliquée est $\beta\Rightarrow^n\Phi,\gamma,\Delta\Rightarrow^1\Phi,\Gamma,\Delta$, alors par hypothèse de récurrence, on peut utiliser la coupure\vspace{0.5cm}

\AxiomC{$\brokenvert$}
\noLine
\UnaryInfC{$\Phi,\gamma,\Delta\vdash\beta$}
\AxiomC{}
\RightLabel{(pAx.$_I$)}
\UnaryInfC{$\Gamma\vdash\gamma$}
\RightLabel{(CUT)}
\BinaryInfC{$\Phi,\Gamma,\Delta,\vdash\beta$}
\DisplayProof\vspace{0.5cm}

D'où le résultat souhaité pour $\beta=s$, puisque les symboles terminaux sont $\m{T}$ et qu'il n'y a donc pas de variable aux feuilles. Remarquons que la dérivations (le $\lambda$-terme) créée est bien identique à la première, c'est juste l'ordre des coupures (correspondant aux substitutions) qui diffère.
\end{itemize}
\end{proof}



\section{Exemples 1 et 2, nécessitant une infinité d'axiomes propres}
\label{an:G_1}

Dès l'ordre 3 pour la grammaire $\m{G}_1$ de lexèmes typés $A : (s/r)/s, B : s/(s/r)$ et $C : s$, pour tout $n\in\mathbb{N}$, le séquent $B^nA^nC\vdash s$ est dérivable dans $\m{S}_{IE}$ et demande $n$ introductions imbriquées.
Un exemple pour $n=2$ :\vspace{0.5cm}
\parindent=-0.5cm

{\footnotesize
\AxiomC{$B\vdash B:s/(s/r)$}
\AxiomC{$B\vdash B:s/(s/r)$}
\AxiomC{$A\vdash A:(s/r)/s$}
\AxiomC{$A\vdash A:(s/r)/s$}
\AxiomC{$C\vdash C:s$}
\RightLabel{(/E)}
\BinaryInfC{$AC\vdash \app{A}{C}:s/r$}
\AxiomC{$\cB{x\!:r}\vdash \cB{x\!:r}$}
\RightLabel{(/E)}
\BinaryInfC{$AC,x:r\vdash \app{\app{A}{C}}{x}:s$}
\RightLabel{(/E)}
\BinaryInfC{$AAC,x:r\vdash \app{A}{\app{\app{A}{C}}{x}}:s/r$}
\AxiomC{$\cG{y\!:r}\vdash\cG{y\!:r}$}
\RightLabel{(/E)}
\BinaryInfC{$AAC,x:r,\cG{y\!:r}\vdash \app{\app{A}{\app{\app{A}{C}}{x}}}{y}:s$}
\RightLabel{(/I)}
\UnaryInfC{$AAC,x:r\vdash \uabs{y}{\app{\app{A}{\app{\app{A}{C}}{x}}}{y}}:s/\cG{r}$}
\RightLabel{(/E)}
\BinaryInfC{$BAAC,\cB{x\!:r}\vdash \app{B}{\uabs{y}{\app{\app{A}{\app{\app{A}{C}}{x}}}{y}}}:s$}
\RightLabel{(/I)}
\UnaryInfC{$BAAC\vdash \uabs{x}{\app{B}{\uabs{y}{\app{\app{A}{\app{\app{A}{C}}{x}}}{y}}}}:s/\cB{r}$}
\RightLabel{(/E)}
\BinaryInfC{$BBAAC\vdash \app{B}{\uabs{x}{\app{B}{\uabs{y}{\app{\app{A}{\app{\app{A}{C}}{x}}}{y}}}}}:s$}
\DisplayProof\vspace{0.5cm}
\parindent=1.5em
}

Plus généralement on peut dériver
\[ \app{B}{\uabs{x_1}{\app{...B}{\uabs{x_n}{\app{\app{A...A}{\app{\app{A}{C}}{x_1}}{...x_n}}}}}} \]

Pour mieux visualiser la suppression des introductions imbriquées, voici la traduction de cet exemple dans $\m{S}_{IC}$, après la suppression d'une introduction, puis après celle de la deuxième (donc dans $\m{S}_C$) en figure [\ref{fig:der_G_1}]. On a automatiquement $\beta$-réduit pour obtenir finalement le même $\lambda$-terme. De manière générale, la preuve dans $\m{S}_C$ conserve bien la structure de dérivation à $\beta$-équivalence près.

\begin{figure}[p] \vspace{-1cm}\centering

\rotatebox{-90}{\footnotesize
\AxiomC{$B,z_1:s/r\vdash \app{B}{z_1}:s$}
	\AxiomC{$B,z_1':(s/r)/r\vdash \uabs{x}{\app{B}{\app{z_1'}{x}}}:s/r$}
		\AxiomC{$A,z_1':s/r\vdash \uabs{x'}{\uabs{y}{\app{\app{A}{\app{z_1'}{x'}}}{y}}}:(s/r)/r$}
			\AxiomC{$A,z_1:s\vdash \uabs{x''}{\app{\app{A}{z_1}}{x''}}:s/r$}
			\AxiomC{$C\vdash C:s$}
			\RightLabel{(CUT)}
			\BinaryInfC{$AC\vdash \uabs{x''}{\app{\app{A}{C}}{x''}}:s/r$}
		\RightLabel{(CUT)}
		\BinaryInfC{$AAC\vdash \uabs{x'}{\uabs{y}{\app{\app{A}{\app{\app{A}{C}}{x'}}}{y}}}:(s/r)/r$}
	\RightLabel{(CUT)}
	\BinaryInfC{$BAAC\vdash \uabs{x}{\app{B}{\uabs{y}{\app{\app{A}{\app{\app{A}{C}}{x}}}{y}}}}:s/r$}
\RightLabel{(CUT)}
\BinaryInfC{$BBAAC\vdash \app{B}{\uabs{x}{\app{B}{\uabs{y}{\app{\app{A}{\app{\app{A}{C}}{x}}}{y}}}}}:s$}
\DisplayProof
}\hspace{2cm}
\rotatebox{-90}{\footnotesize
\AxiomC{$B,z_1:s/r\vdash \app{B}{z_1}:s$}
\AxiomC{$B,z_1:s/r\vdash \app{B}{z_1}:s$}
\AxiomC{$A,z_1:s\vdash \uabs{y}{\app{\app{A}{z_1}}{y}}:s/r$}
\AxiomC{$A,z_1:s,z_2:r\vdash \app{\app{A}{z_1}}{z_2}:s$}
\AxiomC{$C\vdash C:s$}
\RightLabel{(CUT)}
\BinaryInfC{$AC,z_2:r\vdash \app{\app{A}{C}}{z_2}:s$}
\AxiomC{$\cB{x\!:r}\vdash \cB{x\!:r}$}
\RightLabel{(CUT)}
\BinaryInfC{$AC,x:r\vdash \app{\app{A}{C}}{x}:s$}
\RightLabel{(CUT)}
\BinaryInfC{$AAC,x:r\vdash \uabs{y}{\app{\app{A}{\app{\app{A}{C}}{x}}}{y}}:s/r$}
\RightLabel{(CUT)}
\BinaryInfC{$BAAC,\cB{x\!:r}\vdash \app{B}{\uabs{y}{\app{\app{A}{\app{\app{A}{C}}{x}}}{y}}}:s$}
\RightLabel{(/I)}
\UnaryInfC{$BAAC\vdash \uabs{x}{\app{B}{\uabs{y}{\app{\app{A}{\app{\app{A}{C}}{x}}}{y}}}}:s/\cB{r}$}
\RightLabel{(CUT)}
\BinaryInfC{$BBAAC\vdash \app{B}{\uabs{x}{\app{B}{\uabs{y}{\app{\app{A}{\app{\app{A}{C}}{x}}}{y}}}}}:s$}
\DisplayProof
}\hspace{2cm}
\rotatebox{-90}{\footnotesize
\AxiomC{$B,z_1:s/r\vdash \app{B}{z_1}:s$}
\AxiomC{$B,z_1:s/r\vdash \app{B}{z_1}:s$}
\AxiomC{$A,z_1:s,z_2:r\vdash \app{\app{A}{z_1}}{z_2}:s$}
\AxiomC{$A,z_1:s,z_2:r\vdash \app{\app{A}{z_1}}{z_2}:s$}
\AxiomC{$C\vdash C:s$}
\RightLabel{(CUT)}
\BinaryInfC{$AC,z_2:r\vdash \app{\app{A}{C}}{z_2}:s$}
\AxiomC{$\cB{x\!:r}\vdash \cB{x\!:r}$}
\RightLabel{(CUT)}
\BinaryInfC{$AC,x:r\vdash \app{\app{A}{C}}{x}:s$}
\RightLabel{(CUT)}
\BinaryInfC{$AAC,x:r,z_2:r\vdash \app{\app{A}{\app{\app{A}{C}}{x}}}{z_2}:s$}
\AxiomC{$\cG{y\!:r}\vdash\cG{y\!:r}$}
\RightLabel{(CUT)}
\BinaryInfC{$AAC,x:r,\cG{y\!:r}\vdash \app{\app{A}{\app{\app{A}{C}}{x}}}{y}:s$}
\RightLabel{(/I)}
\UnaryInfC{$AAC,x:r\vdash \uabs{y}{\app{\app{A}{\app{\app{A}{C}}{x}}}{y}}:s/\cG{r}$}
\RightLabel{(CUT)}
\BinaryInfC{$BAAC,\cB{x\!:r}\vdash \app{B}{\uabs{y}{\app{\app{A}{\app{\app{A}{C}}{x}}}{y}}}:s$}
\RightLabel{(/I)}
\UnaryInfC{$BAAC\vdash \uabs{x}{\app{B}{\uabs{y}{\app{\app{A}{\app{\app{A}{C}}{x}}}{y}}}}:s/\cB{r}$}
\RightLabel{(CUT)}
\BinaryInfC{$BBAAC\vdash \app{B}{\uabs{x}{\app{B}{\uabs{y}{\app{\app{A}{\app{\app{A}{C}}{x}}}{y}}}}}:s$}
\DisplayProof
}\hspace{1cm}
\caption{Dérivation de l'exemple 1 dans $\m{S}_{IC}$ tout à droite, intermédiaire (dans $\m{S}_C^1$) avec une seule introduction supprimée au milieu, et dans $\m{S}_C$ tout à gauche\label{fig:der_G_1}}
\end{figure}

De même à l'ordre 4 pour $\m{G}_2$ définie par $A : s/(s/(s\backslash s))$ et $B : s$, alors pour tout $n\in\mathbb{N}$, $A^nB\vdash s$ est dérivable dans $\m{S}_{IE}$, et demande $n$ introductions imbriquées. Un preuve pour $n=2$ a pour $\lambda$-terme $\app{A}{\uabs{x}{\app{A}{\uabs{y}{\app{x}{\app{y}{B}}}}}}$, où une variable arrive en tête.

\end{document}